\documentclass{article} \usepackage{iclr2025_conference,times}
\iclrfinalcopy \usepackage[english]{babel} 		

\usepackage{amsfonts}				\usepackage{amsmath}				\usepackage{siunitx}  				\DeclareSIUnit{\hours}{hours}
\usepackage{xfrac}					\usepackage{bm}						\usepackage{mathtools}
\usepackage{mathabx}				\usepackage{nicefrac}

\usepackage{xspace}
\usepackage[utf8]{inputenc} \usepackage[T1]{fontenc}    \usepackage{microtype}      \usepackage{xcolor}         

\usepackage{hyphenat}

\usepackage{graphicx}
\usepackage[labelfont=bf]{caption}
\usepackage{subcaption}
\usepackage{tikz}
\usepackage{pifont}\usepackage[export]{adjustbox}

\usepackage{enumitem}

\usepackage{booktabs}               \usepackage{multirow}               \usepackage{soul}                   \usepackage{changepage,threeparttable} \usepackage{tabularx}               \usepackage{multirow}               \usepackage{etoolbox}               \usepackage{makecell}               
\usepackage{colortbl}               \newcolumntype{g}{>{\columncolor{TUgray_vlight}}l}  
\newrobustcmd{\B}{\bfseries}

\definecolor{mydarkblue}{rgb}{0,0.08,0.45} \usepackage[
    colorlinks=true,
    linkcolor=mydarkblue,
    citecolor=mydarkblue,
    filecolor=mydarkblue,
    urlcolor=mydarkblue,
]{hyperref}       		\usepackage{url}
\usepackage[nameinlink,capitalize,noabbrev]{cleveref}

\let\oldFootnote\footnote
\newcommand\nextToken\relax

\renewcommand\footnote[1]{\oldFootnote{#1}\futurelet\nextToken\isFootnote}

\newcommand\isFootnote{\ifx\footnote\nextToken\textsuperscript{,}\fi}

\newcommand{\iclrfinalfootnote}[1]{
  \ificlrfinal
    \footnote{#1}\fi
}

\makeatletter
\newcommand{\iclrfinaltext}[1]{\@bsphack
  \ificlrfinal
    #1\fi
  \@esphack
}
\makeatother

\newcommand*{\eg}{e.g.\@\xspace}
\newcommand*{\ie}{i.e.\@\xspace}

\usepackage[super]{nth}

\newcommand{\dl}{deep learning\xspace}

\newcommand{\nn}{neural network\xspace}
\newcommand{\nns}{neural networks\xspace}

\newcommand{\dataset}{dataset\xspace}
\newcommand{\datasets}{datasets\xspace}

\newcommand{\runtime}{runtime\xspace}
\newcommand{\runtimes}{runtimes\xspace}
\newcommand{\Runtime}{Runtime\xspace}
\newcommand{\heldout}{held-out\xspace}

\newcommand{\ta}{training algorithm\xspace}
\newcommand{\tas}{training algorithms\xspace}

\newcommand{\wl}{workload\xspace}
\newcommand{\wls}{workloads\xspace}
\newcommand{\Wl}{Workload\xspace}
\newcommand{\Wls}{Workloads\xspace}
\newcommand{\hp}{hyperparameter\xspace}

\newcommand{\hps}{hyperparameters\xspace}

\newcommand{\algoperf}{\textsc{AlgoPerf}\xspace}
\newcommand{\algoperfTA}{\textsc{AlgoPerf: Training Algorithms}\xspace}
\newcommand{\benchmarkinghardware}{8$\times$NVIDIA V100 GPUs\xspace}              

\newcommand{\adam}{\textsc{\mbox{Adam}}\xspace}
\newcommand{\adamw}{\textsc{\mbox{AdamW}}\xspace}
\newcommand{\sgd}{\textsc{\mbox{SGD}}\xspace}

\newcommand{\nadamw}{\textsc{\mbox{NadamW}}\xspace}

\newcommand{\baseline}{\textsc{Baseline}\xspace}
\newcommand{\sfadam}{\textsc{Schedule Free AdamW}\xspace}
\newcommand{\nadamwseq}{\textsc{NadamW Sequential}\xspace}
\newcommand{\sinvnum}{\textsc{Sinv6 75}\xspace}
\newcommand{\sinv}{\textsc{Sinv6}\xspace}
\newcommand{\adamg}{\textsc{AdamG}\xspace}
\newcommand{\shampoosub}{\textsc{PyTorch Distributed Shampoo}\xspace}
\newcommand{\shampoosubshort}{\textsc{PyTorch Distr. Shampoo}\xspace}
\newcommand{\generalizedadam}{\textsc{Generalized Adam}\xspace}
\newcommand{\cycliclr}{\textsc{Cyclic LR}\xspace}
\newcommand{\nadamp}{\textsc{NadamP}\xspace}
\newcommand{\caspr}{\textsc{CASPR Adaptive}\xspace}
\newcommand{\amos}{\textsc{Amos}\xspace}
\newcommand{\lawaq}{\textsc{LAWA Queue}\xspace}
\newcommand{\lawaema}{\textsc{LAWA EMA}\xspace}
\newcommand{\sfprodigy}{\textsc{Schedule Free Prodigy}\xspace}

\newcommand{\legendlinewidth}{2pt}
\newcommand{\linelength}{2em}

\newcommand{\lineraiseheight}{-0.6ex}

\DeclareRobustCommand{\shampoosubline}{\begin{tikzpicture}[baseline=\lineraiseheight]
    \draw[shampoosubcol,solid,line width=\legendlinewidth, opacity=0.8] (0,0) -- (\linelength,0) node[midway] {\ding{54}};
  \end{tikzpicture}}

\DeclareRobustCommand{\sfadamline}{\begin{tikzpicture}[baseline=\lineraiseheight]
    \draw[sfadamcol,solid,line width=\legendlinewidth, opacity=0.8] (0,0) -- (\linelength,0) node[midway] {\ding{108}};
  \end{tikzpicture}}

\DeclareRobustCommand{\generalizedadamline}{\begin{tikzpicture}[baseline=\lineraiseheight]
    \draw[generalizedadamcol,solid,line width=\legendlinewidth, opacity=0.8] (0,0) -- (\linelength,0) node[midway] {\ding{108}};
  \end{tikzpicture}}

\DeclareRobustCommand{\cycliclrline}{\begin{tikzpicture}[baseline=\lineraiseheight]
    \draw[cycliclrcol,solid,line width=\legendlinewidth, opacity=0.8] (0,0) -- (\linelength,0) node[midway,yshift=0.5ex] {\ding{116}};
  \end{tikzpicture}}

\DeclareRobustCommand{\nadampline}{\begin{tikzpicture}[baseline=\lineraiseheight]
    \draw[nadampcol,solid,line width=\legendlinewidth, opacity=0.8] (0,0) -- (\linelength,0) node[midway] {\scalebox{1.5}{\ding{169}}};
  \end{tikzpicture}}

\DeclareRobustCommand{\baselineline}{\begin{tikzpicture}[baseline=\lineraiseheight]
    \draw[baselinecol,dashed,line width=\legendlinewidth, opacity=0.8] (0,0) -- (\linelength,0) ;
  \end{tikzpicture}}

\DeclareRobustCommand{\amosline}{\begin{tikzpicture}[baseline=\lineraiseheight]
    \draw[amoscol,solid,line width=\legendlinewidth, opacity=0.8] (0,0) -- (\linelength,0) node[midway] {\ding{54}};
  \end{tikzpicture}}

\DeclareRobustCommand{\casprline}{\begin{tikzpicture}[baseline=\lineraiseheight]
    \draw[casprcol,solid,line width=\legendlinewidth, opacity=0.8] (0,0) -- (\linelength,0) node[midway] {\ding{72}};
  \end{tikzpicture}}

\DeclareRobustCommand{\lawaqline}{\begin{tikzpicture}[baseline=\lineraiseheight]
    \draw[lawaqcol,solid,line width=\legendlinewidth, opacity=0.8] (0,0) -- (\linelength,0) node[midway] {\ding{169}};
  \end{tikzpicture}}

\DeclareRobustCommand{\lawaemaline}{\begin{tikzpicture}[baseline=\lineraiseheight]
    \draw[lawaemacol,solid,line width=\legendlinewidth, opacity=0.8] (0,0) -- (\linelength,0) node[midway,yshift=0.5ex] {\ding{116}};
  \end{tikzpicture}}

\DeclareRobustCommand{\sfprodigyline}{\begin{tikzpicture}[baseline=\lineraiseheight]
    \draw[sfprodigycol,solid,line width=\legendlinewidth, opacity=0.8] (0,0) -- (\linelength,0) ;
  \end{tikzpicture}}

\DeclareRobustCommand{\nadamwseqline}{\begin{tikzpicture}[baseline=\lineraiseheight]
    \draw[nadamwseqcol,solid,line width=\legendlinewidth, opacity=0.8] (0,0) -- (\linelength,0) node[midway] {\scalebox{1.5}{\ding{169}}};
  \end{tikzpicture}}

\DeclareRobustCommand{\sinvnumline}{\begin{tikzpicture}[baseline=\lineraiseheight]
    \draw[sinvnumcol,solid,line width=\legendlinewidth, opacity=0.8] (0,0) -- (\linelength,0) node[midway] {\ding{54}};
  \end{tikzpicture}}

\DeclareRobustCommand{\sinvline}{\begin{tikzpicture}[baseline=\lineraiseheight]
    \draw[sinvcol,solid,line width=\legendlinewidth, opacity=0.8] (0,0) -- (\linelength,0) node[midway,yshift=0.5ex] {\ding{116}};
  \end{tikzpicture}}

\DeclareRobustCommand{\adamgline}{\begin{tikzpicture}[baseline=\lineraiseheight]
    \draw[adamgcol,solid,line width=\legendlinewidth, opacity=0.8] (0,0) -- (\linelength,0) node[midway] {\ding{72}};
  \end{tikzpicture}}

\newcommand{\imagenet}{\textsc{ImageNet}\xspace}
\newcommand{\wmt}{\textsc{WMT}\xspace}
\newcommand{\librispeech}{\textsc{LibriSpeech}\xspace}
\newcommand{\criteo}{\textsc{Criteo 1TB}\xspace}
\newcommand{\ogbg}{\textsc{OGBG}\xspace}
\newcommand{\fastmri}{\textsc{fastMRI}\xspace}

\newcommand{\resnetfifty}{\textsc{ResNet-50}\xspace}
\newcommand{\resnet}{\textsc{ResNet}\xspace}

\newcommand{\vit}{\textsc{ViT}\xspace}

\newcommand{\transformer}{\textsc{Transformer}\xspace}
\newcommand{\deepspeech}{\textsc{DeepSpeech}\xspace}
\newcommand{\conformer}{\textsc{Conformer}\xspace}
\newcommand{\dlrmsmall}{\textsc{DLRMsmall}\xspace}
\newcommand{\gnn}{\textsc{GNN}\xspace}
\newcommand{\unet}{\textsc{U-Net}\xspace}

\newcommand{\python}{\textsc{Python}\xspace}

\newcommand{\pytorch}{\textsc{PyTorch}\xspace}
\newcommand{\jax}{\textsc{JAX}\xspace}
\newcommand{\tensorflow}{\textsc{TensorFlow}\xspace}
 
\usepackage{amsmath,amsfonts,bm}

\def\eqref#1{equation~\ref{#1}}

\def\1{\bm{1}}

\DeclareMathAlphabet{\mathsfit}{\encodingdefault}{\sfdefault}{m}{sl}
\SetMathAlphabet{\mathsfit}{bold}{\encodingdefault}{\sfdefault}{bx}{n}

\definecolor{TUred}{RGB}{165,30,55}
\definecolor{TUdark}{RGB}{50,65,75}
\colorlet{TUdark_light}{TUdark!80}
\definecolor{TUgold}{RGB}{180,160,105}

\definecolor{TUgray}{RGB}{185,184,188}
\colorlet{TUgray_light}{TUgray!80}
\colorlet{TUgray_vlight}{TUgray!40}

\definecolor{TUdarkblue}{RGB}{65,90,140}
\definecolor{TUblue}{RGB}{0,105,170}
\definecolor{TUlightblue}{RGB}{80,170,200}

\definecolor{TUlightgreen}{RGB}{125,165,75}
\definecolor{TUgreen}{RGB}{125,165,75}
\definecolor{TUdarkgreen}{RGB}{50,110,30}

\definecolor{TUlightred}{RGB}{200,80,60}
\definecolor{TUpurple}{RGB}{175,110,150}
\definecolor{TUorange}{RGB}{210,150,0}

\definecolor{SNSblue}{rgb}{0.1216, 0.4666, 0.7059}
\definecolor{SNSorange}{rgb}{1.0, 0.4980, 0.0549}
\definecolor{SNSgreen}{rgb}{0.1725, 0.6274, 0.1725}
\definecolor{SNSred}{rgb}{0.84, 0.15, 0.16}
\definecolor{SNSpurple}{rgb}{0.58, 0.40, 0.74}

\definecolor{SNSorange_shaded}{HTML}{ffcea3}
\definecolor{SNSblue_shaded}{HTML}{8ebad9}
\definecolor{SNSgreen_shaded}{HTML}{cae7ca}
\definecolor{SNSred_shaded}{HTML}{ea9293}

\definecolor{MPLred_shaded}{HTML}{df735b}
\definecolor{MPLblue_shaded}{HTML}{3885bc}

\definecolor{PlotRed}{rgb}{0.77, 0.31, 0.32}
\definecolor{PlotBlue}{rgb}{0.30, 0.45, 0.69}

\definecolor{shampoosubcol}{rgb}{0.09, 0.745, 0.81}
\definecolor{sfadamcol}{rgb}{0.12, 0.467, 0.706}
\definecolor{generalizedadamcol}{rgb}{1.0, 0.5, 0.055}
\definecolor{cycliclrcol}{rgb}{0.1725, 0.6274, 0.1725}
\definecolor{nadampcol}{rgb}{0.8392, 0.1529, 0.1569}
\definecolor{baselinecol}{HTML}{32414b}
\definecolor{amoscol}{rgb}{0.737, 0.741, 0.133}
\definecolor{casprcol}{rgb}{0.549, 0.337, 0.294}
\definecolor{lawaqcol}{rgb}{0.58 0.40, 0.74}
\definecolor{lawaemacol}{rgb}{0.82, 0.73, 1.0}
\definecolor{sfprodigycol}{HTML}{696969}
\definecolor{nadamwseqcol}{rgb}{1.0, 0.498, 0.055}
\definecolor{sinvnumcol}{rgb}{0.58, 0.404, 0.74}
\definecolor{sinvcol}{rgb}{0.8392, 0.153, 0.157}
\definecolor{adamgcol}{rgb}{0.1725, 0.6274, 0.1725}
  
\title{Accelerating neural network training:\\An analysis of the AlgoPerf competition}

\author{\textbf{Priya Kasimbeg}$^1$\thanks{Equal contributions.} \quad
    \textbf{Frank Schneider}$^2$\footnotemark[\value{footnote}] \quad
    \textbf{Runa Eschenhagen}$^3$ \quad
    \textbf{Juhan Bae}$^{4,5}$ \quad \\
    \textbf{Chandramouli Shama Sastry}$^{4,6}$ \quad
    \textbf{Mark Saroufim}$^{7}$ \quad
    \textbf{Boyuan Feng}$^{7}$ \quad
    \textbf{Less Wright}$^{7}$ \quad \\
    \textbf{Edward Z. Yang}$^{7}$  \quad 
    \textbf{Zachary Nado}$^1$ \quad
    \textbf{Sourabh Medapati}$^1$ \quad \\
    \textbf{Philipp Hennig}$^2$ \quad
    \textbf{Mike Rabbat}$^7$ \quad 
    \textbf{George E.~Dahl}$^1$\thanks{Corresponding author gdahl@google.com.} \\
     $^1$Google DeepMind \quad
     $^2$University of Tübingen \quad
     $^3$University of Cambridge \quad \\
     $^4$Vector Institute \quad
     $^5$University of Toronto \quad
     $^6$Dalhousie University \quad
     $^7$Meta \quad
}

\begin{document}
\maketitle

\begin{abstract}
    The goal of the \algoperfTA competition is to evaluate practical speed-ups in \nn training achieved solely by improving the underlying \tas.
    In the external tuning ruleset, submissions must provide \wl-agnostic \hp search spaces, while in the self-tuning ruleset they must be completely hyperparameter-free.
    In both rulesets, submissions are compared on time-to-result across multiple \dl \wls, training on fixed hardware.
    This paper presents the inaugural \algoperf competition's results, which drew $18$ diverse submissions from $10$ teams.
    Our investigation reveals several key findings:
    (1) The winning submission in the external tuning ruleset, using \textsc{Distributed Shampoo}, demonstrates the effectiveness of non-diagonal preconditioning over popular methods like \adam, even when compared on wall-clock runtime.
    (2) The winning submission in the self-tuning ruleset, based on the \sfadam algorithm, demonstrates a new level of effectiveness for completely hyperparameter-free training algorithms.
    (3) The top-scoring submissions were surprisingly robust to \wl changes.
    We also discuss the engineering challenges encountered in ensuring a fair comparison between different \tas.
    These results highlight both the significant progress so far, and the considerable room for further improvements.
\end{abstract}

\section{Introduction}
\label{sec:intro}
Deep \nns are powerful models that excel in tasks such as image recognition, natural language processing, and speech recognition.
However, training these models often requires significant computational resources as well as careful (and sometimes brittle) training recipes, including meticulous \hp tuning.
A big practical issue in the usability of \emph{\tas}, for instance, is that many choices are still left to the practitioner.
How should the learning rate be tuned? In what range? Using what schedule?
These are very crucial decisions that can make or break the training process and, critically, determine which methods perform best.
Despite training algorithms being such a fundamental part of the deep learning pipeline, the community has been unable to identify which training methods are the state of the art.
Previous empirical comparisons have suffered from a number of issues, \eg, weak baselines, not fully accounting for hyperparameter tuning, or failing to properly control for potential confounding factors like model architecture changes.
\citet{Dahl2023AlgoPerf} proposed the \algoperfTA benchmark (\cref{sec:methods}) to measure speed-ups in neural net training due to algorithmic improvements, while addressing the aforementioned issues.
This competitive, time-to-result benchmark uses multiple realistic \dl \wls on fixed hardware, allowing submitters to innovate solely on the \tas.
It allows submissions under two different \hp tuning rulesets: an \emph{external tuning ruleset} that scores submissions based on the best result from a handful of \hp configurations, and a \emph{self-tuning ruleset} without \hps that counts any time a submission spends tuning as part of the training time.

In this paper, we present the results of the inaugural \algoperfTA competition, a competition open to the entire machine learning community, based on the \algoperf benchmark proposed in \citet{Dahl2023AlgoPerf}.
Our analysis of $\approx\!\num{4000}$ training runs reveals several key findings on accelerating neural net training through improved \tas:
\begin{itemize}[itemsep=0pt,leftmargin=1em]
    \item The winning submission in the external tuning ruleset, based on \textsc{Distributed Shampoo} \citep{anil2020shampoo,shi2023distributeddataparallelpytorchimplementation}, demonstrates that non-diagonal preconditioning methods can outperform currently popular diagonal methods, such as \adam \citep{Kingma2015}, in terms of wall-clock training time.
    Across eight \dl \wls, this submission achieved on average a $\approx\!\!28\%$ faster model training compared to the baseline, which used \nadamw \citep{Dozat2016,Loshchilov2019} (\cref{sec:results}).
    \item In the self-tuning ruleset, the winning submission based on \sfadam \citep{defazio2024road} was the only entry surpassing the baseline, providing $\approx\!\!8\%$ faster average training.
    It was also $\approx\!\!10\%$ faster than the \emph{external} tuning baseline across the seven base benchmark \wls both algorithms trained successfully, without any \hps or parallel tuning (\cref{sec:results}).\footnote{Note, the two speed-ups for \sfadam are computed across different sets of \wls.}
    This result establishes a new state-of-the-art for \hp-free \tas and highlights the exciting potential of fully automated \nn training.
    \item The top-scoring submissions are characterized by their consistent performance across \wls, including robustness to minor workload modifications, \eg changes to activation functions or normalization layers (\cref{tab:held_out_workloads}).
    This suggests that, at least in a competitive evaluation context, achieving consistent performance across \wls is a major challenge for \tas operating under restricted runtime and tuning budgets.
\end{itemize}

Building the training harness and software for the \algoperfTA competition to enable fair and meaningful comparisons between \tas, especially across the \dl frameworks \jax \citep{Bradbury2018} and \pytorch \citep{Paszke2019}, required substantial engineering effort.
\cref{sec:engineering} discusses these engineering challenges, while \cref{sec:lessons} highlights lessons learned as well as opportunities to improve future iterations of the benchmark.

\section{Summary of benchmarking methodology}
\label{sec:methods}

The \algoperfTA benchmark evaluates the effectiveness of \tas by measuring how quickly they can achieve specific evaluation metric goal values across various realistic \dl \wls.
These per-\wl, time-to-result measurements are performed on a fixed hardware configuration, and account for all required \wl-specific tuning. The final benchmark score aggregates across \wls to identify more efficient general-purpose \tas for \dl.
Below, we briefly summarize the \algoperf benchmark by explaining key terms.
For a more detailed discussion of the benchmark's motivation, description, and justification see  \citet{Dahl2023AlgoPerf}\iclrfinaltext{ and the competition rules\footnote{\href{https://github.com/mlcommons/algorithmic-efficiency/blob/main/docs/.old/COMPETITION_RULES.md}{\nolinkurl{github.com/mlcommons/algorithmic-efficiency/[...]/COMPETITION_RULES.md}}} \& documentation\footnote{\href{https://github.com/mlcommons/algorithmic-efficiency/blob/main/docs/DOCUMENTATION.md}{\nolinkurl{github.com/mlcommons/algorithmic-efficiency/[...]/DOCUMENTATION.md}}}}.
For the \algoperf competition, we solicited submissions from the entire machine learning community, and made several modifications to the benchmark as it was initially proposed by \citet{Dahl2023AlgoPerf}, which are summarized in \cref{sec:methods_mods}.

\textbf{Workloads.} The benchmark features multiple neural network training tasks, called \emph{\wls}, each consisting of a dataset, model, loss function, target metric, validation target and \runtime budget.
The submissions' objective is to train these \wls as quickly as possible; if the target is not reached within the \runtime budget, the run receives an infinite score.
Designed to reflect real-world \dl training scenarios, the \wls cover several key domains.
The benchmark includes two types of \wls: \emph{fixed base \wls} (\cref{tab:workloads}) directly affect the benchmark score, and \emph{held-out \wl variants} (\cref{tab:held_out_workloads}) discourage submissions to overfit the benchmark's fixed \wls and ensure robustness to natural \wl changes.
While held-out \wls do not contribute directly to the benchmark score, failure to train a \heldout \wl quickly enough will invalidate the submission's score for the corresponding fixed base \wl.

\textbf{Submissions.} Submitted \tas must adhere to the fixed \algoperf API \citep[Sec.~4.2]{Dahl2023AlgoPerf} and are limited to four submission functions:
(1) \texttt{update\_params} is responsible for modifying the network's parameters during training and typically involves optimization algorithms, such as \sgd, \adam, or a custom method.
(2) \texttt{init\_optimizer\_state} allows the creating of the \ta's internal state, \eg to define learning rate schedules.
(3) \texttt{data\_selection} allows control of how batches are constructed, to use techniques such as curriculum learning or data echoing \citep{Choi2020}.
(4) \texttt{get\_batch\_size} defines a batch size for each \wl, \eg, participants can predetermine the largest batch size fitting in the competition hardware's memory.
For the external tuning ruleset (see below), participants can also provide a \wl-agnostic search space for \hps.
This limited API isolates training speed-ups resulting from improvements to the \ta itself, rather than pipeline optimizations or model changes.
The API also ensures that submissions can be applied to generic \dl \wls by being fully specified without any \wl-specific behavior (apart from the batch size).

\textbf{Tuning rulesets.} Submissions compete under two distinct rulesets governing \hp tuning.
The \emph{external tuning ruleset} simulates \hp tuning with limited parallel resources.
In this ruleset, \hps are tuned using five independent \emph{trials}, with \hp configurations sampled via quasirandom search \citep{Bousquet2017} from the submission's defined search space,\footnote{Submissions can alternatively provide a list of five \hp configurations that will be sampled without replacement.} and are scored based on the \runtime of the trial that achieves the validation target the fastest.
To produce a more consistent final score, the tuning process is repeated five times across five different tuning \emph{studies}, where each study receives a different random seed for the \wl's model and data initialization. The final \emph{\wl score} used in the benchmark's scoring procedure is the median of the best training time from each of the five studies.
The \emph{self-tuning ruleset} simulates fully automated \hp tuning during training on a single machine.
This includes submissions that use the same \hps across all \wls (\eg \adamw with defaults for all \hps including regularization) or those that perform inner-loop tuning during the training run.
Anticipating that they may require more time to reach the target, self-tuning submissions have three times the \runtime budget of external-tuning submissions (see \cref{tab:workloads}).
Similar to the external tuning ruleset, five studies are conducted (although each self-tuning study consists of a single trial), and the median \runtime across all studies determines the \wl score.

\textbf{Benchmark score.} A submission's \emph{benchmark score} is based on its individual \wl scores relative to those of other submissions, aggregated using performance profiles \citep[Sec.~4.5]{Dolan2002, Dahl2023AlgoPerf}.
A performance profile plots the fraction of \wls where a submission trains successfully (achieves the target performance) within a factor of $\tau$ of the time required by the per-\wl fastest submission, for different values of $1 \!\leq\! \tau \!\leq\! \tau_{\text{max}}$ (see \cref{fig:perf_profiles,fig:scores_max_tau}).
For instance, the height of submission $s$'s performance profile at $\tau\!=\!1.5$ is the fraction of \wls where $s$ trains successfully and within $1.5\!\times$ the wall-clock time the best competitor requires on that \wl.
The final scalar \emph{benchmark scores} (\cref{tab:leaderboard_et,tab:leaderboard_st}) are the normalized area under the performance profiles, with $1.0$ corresponding to a submission that was faster than all competitors on all \wls.

\textbf{Computational costs.} To score all competition submissions, we conducted \num{3850} and \num{420} runs in the external tuning and self-tuning ruleset respectively.
On average, scoring an external submission required $\approx\!\SI{3469}{\hours}$, and $\approx\!\SI{1847}{\hours}$ for a self-tuning submission, totaling $\approx\!\SI{49240}{\hours}$ on the competition hardware (\benchmarkinghardware) (see \cref{app:costs}).

\section{Results}
\label{sec:results}

\begin{figure}
  \centering
  \begin{minipage}[t]{0.36\textwidth}
    \subfloat[t][\textbf{External tuning} leaderboard\label{tab:leaderboard_et}]{
      \scriptsize
    {\renewcommand{\arraystretch}{1.1}
  \setlength{\tabcolsep}{4pt}
  \hyphenpenalty=10000\exhyphenpenalty=10000
  \begin{tabularx}{\textwidth}{>{\raggedright\arraybackslash}XcS[table-format=1.4]}
    \toprule
    \textbf{Submission} & \textbf{Line} & \textbf{Score} \\ \midrule
    \shampoosub & \shampoosubline & 0.7784 \\
    \sfadam & \sfadamline & 0.7077 \\
    \generalizedadam & \generalizedadamline & 0.6383 \\
    \cycliclr & \cycliclrline & 0.6301 \\
    \nadamp & \nadampline & 0.5909 \\
    \rowcolor{TUgray_light}\baseline & \baselineline & 0.5707 \\
    \amos & \amosline & 0.4918 \\
    \caspr & \casprline & 0.4722 \\
    \lawaq & \lawaqline & 0.3699 \\
    \lawaema & \lawaemaline & 0.3384 \\
    \sfprodigy & \sfprodigyline & 0 \\
    \bottomrule
  \end{tabularx}
\vspace{0pt}}
 }
  \end{minipage}\hspace{1.5em}\begin{minipage}[t]{0.6\textwidth}
    \subfloat[t][\textbf{External tuning} performance profiles]{
      \includegraphics[width=\linewidth,valign=t]{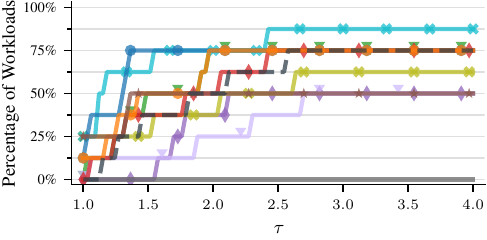}
      \label{fig:pp_external_tuning}
    }
  \end{minipage}\\[1em]
  \begin{minipage}[t]{0.36\textwidth}
    \subfloat[t][\textbf{Self-tuning} leaderboard\label{tab:leaderboard_st}]{
      \scriptsize
    {\renewcommand{\arraystretch}{1.25}
  \setlength{\tabcolsep}{4pt}
  \hyphenpenalty=10000\exhyphenpenalty=10000
  \begin{tabularx}{\textwidth}{>{\raggedright\arraybackslash}XcS[table-format=1.4]}
    \toprule
    \textbf{Submission} & \textbf{Line} & \textbf{Score} \\ \midrule
    \sfadam & \sfadamline & 0.8542 \\
    \rowcolor{TUgray_light}\baseline & \baselineline & 0.8194 \\
    \nadamwseq & \nadamwseqline & 0.3308 \\
    \sinvnum & \sinvnumline & 0.1420 \\
    \sinv & \sinvline & 0.0903 \\
    \adamg & \adamgline & 0 \\
    \bottomrule
  \end{tabularx}
\vspace{3.33\baselineskip}
}
 }
  \end{minipage}\hspace{1.5em}\begin{minipage}[t]{0.6\textwidth}
    \subfloat[t][\textbf{Self-tuning} performance profiles]{
      \includegraphics[width=\linewidth,valign=t]{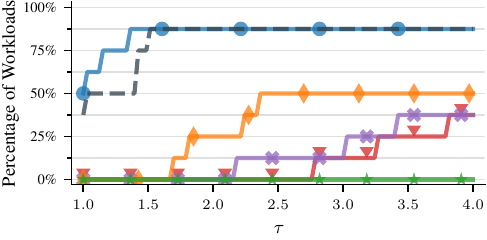}
      \label{fig:pp_self_tuning}
    }
  \end{minipage}
  \caption{\textbf{\algoperf competition leaderboard \& performance profiles} for all submissions to the external (\textit{top}) and self-tuning (\textit{bottom}) ruleset. The leaderboards (\subref{tab:leaderboard_et}, \subref{tab:leaderboard_st}) are ranked by the submissions' benchmark scores, rounded to four significant digits. Higher scores indicate faster training. Note, scores are \emph{not} comparable between rulesets. In the performance profiles (\subref{fig:pp_external_tuning}, \subref{fig:pp_self_tuning}), each line represents a submission. A step at $\tau$ indicates that, for one \wl, this submission reaches the target within $\tau$ times the runtime of the fastest submission for that \wl and ruleset.}
  \label{fig:perf_profiles}
\end{figure}

The two \algoperfTA competition leaderboards (\Cref{fig:perf_profiles}) rank submissions in each tuning ruleset by their benchmark score.\footnote{Note that the external-tuning baseline is different from the self-tuning baseline; each was tuned and specialized specifically for its ruleset. Similarly, there are two distinct \sfadam submissions, one for each ruleset. We specify which one we are referring to if the context does not make it clear.}
Although five submissions outperformed the baseline (\cref{tab:submissions_et}) in the external tuning ruleset, only the winning submission was competitive in the self-tuning ruleset.
The remaining self-tuning submissions (\cref{tab:submissions_st}) scored significantly lower than the self-tuning baseline, highlighting the challenge of fully automatic training algorithms that must pay the training time cost of any \wl-specific \hp tuning.
Although there is at least one submission that could successfully train each \wl, no submission reached the target on every \wl (\cref{fig:perf_profiles}; see \cref{sec:results_runtimes} for more details).
This result simultaneously demonstrates the benchmark's feasibility and its difficulty as well as the significant potential for future improvement in \tas.
There were strong submissions in both \pytorch \& \jax, suggesting that, within the benchmark's codebase, workload implementations in both frameworks are, at least to some degree, sufficiently similar in speed and memory usage (see \cref{sec:engineering}).

\subsection{Per-workload runtimes}
\label{sec:results_runtimes}

\Cref{tab:runtimes} provides a detailed overview of the submissions' runtimes across all \wls, normalized by the \emph{external tuning runtime budget}.
\Wls with runtimes deemed infinite for scoring purposes are marked in gray, with corresponding symbols explaining the reason (see the caption of \cref{tab:runtimes}).
The winning submissions in both rulesets excelled by reliably training most \wls rather than being the fastest on every one.
For example, \shampoosub was the fastest on ``only'' $2$ of the $8$ base \wls, while \sfadam led in $4$ of $8$ in the self-tuning ruleset.
We also observed substantial variations in \wl runtimes, even among competitive methods.
\shampoosub, for instance, took over twice as long on \wmt as the fastest submission.
This indicates that there is still ample potential to improve \tas.
In the self-tuning ruleset, \emph{if} the winning submission successfully trained a \wl, it did so within the \emph{external} tuning runtime budget.
This suggests that future benchmarks could significantly reduce the self-tuning runtime budget to save computational costs, perhaps matching the external tuning one.

\begin{table}[!t]
    \caption{\textbf{Normalized submission \runtimes across all \wls.}
    All \runtimes are normalized using the \emph{external tuning} \runtime budget, with the fastest submission per \wl in each ruleset highlighted in bold. \Wl \runtimes considered infinite for scoring are marked in gray, with a (suffix) symbol explaining the reason. \texttt{inf} denotes that a submission did not reach the \wl target within the allowed \runtime budget. \texttt{NaN} indicates an error (such as running out of memory) before any evaluation occurred. A \textdagger\xspace indicates that a \heldout score is ignored due to the submission not reaching the target on the corresponding base workload, while a \textdaggerdbl\xspace indicates that a base \wl score is ignored because the submission did not successfully train the associated \heldout \wl. Runs that are not within $4\times$ the fastest (valid) \wl \runtime are marked with \textasteriskcentered.}
    \label{tab:runtimes}
    \begin{subtable}[c]{\linewidth}
      \centering
        \caption{\textbf{External tuning ruleset}}
        \label{tab:runtimes_et}
        \scriptsize
	    {\renewcommand{\arraystretch}{1.25}\setlength{\tabcolsep}{4pt}\begin{tabularx}{0.99\textwidth}{>{\raggedright\arraybackslash}Xlglglgllgllglg}
\toprule
 & \multicolumn{2}{c}{\criteo} & \multicolumn{2}{c}{\fastmri} & \multicolumn{2}{c}{\resnet} & \vit & \multicolumn{2}{c}{\conformer} & \makecell{\textsc{Deep}\\ \textsc{Speech}} & \multicolumn{2}{c}{\ogbg} & \multicolumn{2}{c}{\wmt} \\
\cmidrule(lr){2-3} \cmidrule(lr){4-5} \cmidrule(lr){6-7} \cmidrule(lr){8-8} \cmidrule(lr){9-10} \cmidrule(lr){11-11} \cmidrule(lr){12-13} \cmidrule(lr){14-15} & Base & H.O. & Base & H.O. & Base & H.O. & Base & Base & H.O. & Base & Base & H.O. & Base & H.O. \\
\midrule
\amos & \textcolor{TUdark_light}{\texttt{inf}} & \textcolor{TUdark_light}{\texttt{inf}} & 0.33 & 0.49 & \textcolor{TUdark_light}{\texttt{inf}} & \textcolor{TUdark_light}{\B 0.55\textsuperscript{\textdagger}} & 0.65 & \B 0.71 & 0.57 & \B 0.57 & \textcolor{TUdark_light}{0.60\textsuperscript{\textasteriskcentered}} & \textcolor{TUdark_light}{0.89\textsuperscript{\textasteriskcentered}} & 0.68 & 0.37 \\
\baseline & 0.94 & 0.08 & 0.23 & 0.51 & \textcolor{TUdark_light}{\texttt{inf}} & \textcolor{TUdark_light}{0.94\textsuperscript{\textdagger}} & 0.91 & 0.90 & 0.83 & 0.65 & \textcolor{TUdark_light}{0.42\textsuperscript{\textdaggerdbl}} & \textcolor{TUdark_light}{0.68\textsuperscript{\textasteriskcentered}} & 0.86 & 0.35 \\
\textsc{CASPR} \newline \textsc{Adaptive} & \textcolor{TUdark_light}{\texttt{NaN}} & \textcolor{TUdark_light}{\texttt{NaN}} & \B 0.13 & \B 0.15 & \textcolor{TUdark_light}{\texttt{inf}} & \textcolor{TUdark_light}{\texttt{inf}} & 0.58 & \textcolor{TUdark_light}{\texttt{inf}} & \textcolor{TUdark_light}{0.59\textsuperscript{\textdagger}} & 0.75 & \B 0.12 & \B 0.12 & \textcolor{TUdark_light}{0.67\textsuperscript{\textdaggerdbl}} & \textcolor{TUdark_light}{\texttt{NaN}} \\
\cycliclr & 0.67 & 0.08 & 0.25 & 0.44 & \textcolor{TUdark_light}{\texttt{inf}} & \textcolor{TUdark_light}{\texttt{inf}} & 0.81 & 0.94 & 0.92 & 0.70 & \textcolor{TUdark_light}{0.38\textsuperscript{\textdaggerdbl}} & \textcolor{TUdark_light}{0.51\textsuperscript{\textasteriskcentered}} & 0.49 & 0.35 \\
\generalizedadam & 0.83 & 0.05 & 0.18 & 0.39 & \B 0.97 & 0.88 & 0.84 & \textcolor{TUdark_light}{\texttt{inf}} & \textcolor{TUdark_light}{0.83\textsuperscript{\textdagger}} & 0.68 & \textcolor{TUdark_light}{0.31\textsuperscript{\textdaggerdbl}} & \textcolor{TUdark_light}{0.64\textsuperscript{\textasteriskcentered}} & 0.63 & 0.33 \\
\lawaema & 0.69 & 0.09 & 0.29 & 0.57 & \textcolor{TUdark_light}{\texttt{inf}} & \textcolor{TUdark_light}{\texttt{inf}} & 0.80 & \textcolor{TUdark_light}{\texttt{inf}} & \textcolor{TUdark_light}{\texttt{inf}} & \textcolor{TUdark_light}{\texttt{inf}} & \textcolor{TUdark_light}{0.57\textsuperscript{\textasteriskcentered}} & \textcolor{TUdark_light}{0.73\textsuperscript{\textasteriskcentered}} & 0.89 & 0.39 \\
\lawaq & \textcolor{TUdark_light}{\texttt{inf}} & \textcolor{TUdark_light}{0.14\textsuperscript{\textdagger}} & 0.22 & 0.55 & \textcolor{TUdark_light}{\texttt{inf}} & \textcolor{TUdark_light}{\texttt{inf}} & 0.66 & \textcolor{TUdark_light}{\texttt{inf}} & \textcolor{TUdark_light}{\texttt{inf}} & \textcolor{TUdark_light}{\texttt{inf}} & 0.25 & 0.24 & 0.56 & 0.22 \\
\nadamp & 0.80 & 0.07 & 0.22 & 0.49 & \textcolor{TUdark_light}{\texttt{inf}} & \textcolor{TUdark_light}{0.90\textsuperscript{\textdagger}} & 0.88 & 0.94 & 0.85 & 0.60 & \textcolor{TUdark_light}{0.43\textsuperscript{\textdaggerdbl}} & \textcolor{TUdark_light}{0.74\textsuperscript{\textasteriskcentered}} & 0.80 & 0.47 \\
\sfadam & 0.67 & 0.05 & 0.13 & 0.41 & \textcolor{TUdark_light}{\texttt{inf}} & \textcolor{TUdark_light}{\texttt{inf}} & 0.57 & 0.92 & \B 0.57 & 0.78 & \textcolor{TUdark_light}{0.29\textsuperscript{\textdaggerdbl}} & \textcolor{TUdark_light}{0.61\textsuperscript{\textasteriskcentered}} & \B 0.33 & \B 0.12 \\
\sfprodigy & \textcolor{TUdark_light}{\texttt{NaN}} & \textcolor{TUdark_light}{\texttt{NaN}} & \textcolor{TUdark_light}{0.21\textsuperscript{\textdaggerdbl}} & \textcolor{TUdark_light}{0.65\textsuperscript{\textasteriskcentered}} & \textcolor{TUdark_light}{\texttt{inf}} & \textcolor{TUdark_light}{\texttt{inf}} & \textcolor{TUdark_light}{\texttt{inf}} & \textcolor{TUdark_light}{\texttt{inf}} & \textcolor{TUdark_light}{\texttt{inf}} & \textcolor{TUdark_light}{\texttt{inf}} & \textcolor{TUdark_light}{0.61\textsuperscript{\textasteriskcentered}} & \textcolor{TUdark_light}{\texttt{inf}} & \textcolor{TUdark_light}{\texttt{inf}} & \textcolor{TUdark_light}{0.40\textsuperscript{\textdagger}} \\
\pytorch \newline \textsc{Distr. Shampoo} & \B 0.65 & \B 0.03 & 0.15 & 0.22 & \textcolor{TUdark_light}{\texttt{inf}} & \textcolor{TUdark_light}{0.93\textsuperscript{\textdagger}} & \B 0.43 & 0.78 & 0.68 & 0.62 & 0.18 & 0.19 & 0.80 & 0.25 \\
\bottomrule
\end{tabularx}}

     \end{subtable}
    \par\bigskip \begin{subtable}[c]{\linewidth}
      \centering
        \caption{\textbf{Self-tuning ruleset}}
        \label{tab:runtimes_st}
    	\scriptsize
    	{\renewcommand{\arraystretch}{1.25}\setlength{\tabcolsep}{4pt}\begin{tabularx}{0.98\textwidth}{>{\raggedright\arraybackslash}Xlglglgllgllglg}
\toprule
 & \multicolumn{2}{c}{\criteo} & \multicolumn{2}{c}{\fastmri} & \multicolumn{2}{c}{\resnet} & \vit & \multicolumn{2}{c}{\conformer} & \makecell{\textsc{Deep}\\ \textsc{Speech}} & \multicolumn{2}{c}{\ogbg} & \multicolumn{2}{c}{\wmt} \\
\cmidrule(lr){2-3} \cmidrule(lr){4-5} \cmidrule(lr){6-7} \cmidrule(lr){8-8} \cmidrule(lr){9-10} \cmidrule(lr){11-11} \cmidrule(lr){12-13} \cmidrule(lr){14-15} & Base & H.O. & Base & H.O. & Base & H.O. & Base & Base & H.O. & Base & Base & H.O. & Base & H.O. \\
\midrule
\adamg & \textcolor{TUdark_light}{\texttt{inf}} & \textcolor{TUdark_light}{\texttt{inf}} & \textcolor{TUdark_light}{\texttt{inf}} & \textcolor{TUdark_light}{\texttt{inf}} & \textcolor{TUdark_light}{\texttt{inf}} & \textcolor{TUdark_light}{\texttt{inf}} & \textcolor{TUdark_light}{\texttt{inf}} & \textcolor{TUdark_light}{\texttt{inf}} & \textcolor{TUdark_light}{\texttt{inf}} & \textcolor{TUdark_light}{\texttt{inf}} & \textcolor{TUdark_light}{\texttt{inf}} & \textcolor{TUdark_light}{\texttt{inf}} & \textcolor{TUdark_light}{\texttt{inf}} & \textcolor{TUdark_light}{\texttt{inf}} \\
\baseline & 0.75 & \B 0.07 & 0.22 & 0.51 & \textcolor{TUdark_light}{\texttt{inf}} & \textcolor{TUdark_light}{\texttt{inf}} & 0.95 & \B 0.94 & 0.92 & \B 0.65 & 0.46 & 0.69 & \B 0.84 & 0.59 \\
\textsc{NadamW}\newline \textsc{Sequential} & \textcolor{TUdark_light}{2.96\textsuperscript{\textdaggerdbl}} & \textcolor{TUdark_light}{0.57\textsuperscript{\textasteriskcentered}} & 0.27 & \B 0.44 & \textcolor{TUdark_light}{\texttt{inf}} & \textcolor{TUdark_light}{\texttt{inf}} & 1.58 & \textcolor{TUdark_light}{\texttt{inf}} & \textcolor{TUdark_light}{1.16\textsuperscript{\textdagger}} & 1.45 & 0.55 & 0.96 & \textcolor{TUdark_light}{2.36\textsuperscript{\textdaggerdbl}} & \textcolor{TUdark_light}{1.57\textsuperscript{\textasteriskcentered}} \\
\sfadam & \B 0.75 & 0.25 & \B 0.15 & 0.58 & \textcolor{TUdark_light}{\texttt{inf}} & \textcolor{TUdark_light}{\texttt{inf}} & \B 0.68 & 0.97 & \B 0.61 & 0.88 & \B 0.32 & \B 0.56 & 0.94 & \B 0.21 \\
\sinv & \textcolor{TUdark_light}{\texttt{NaN}} & \textcolor{TUdark_light}{\texttt{NaN}} & 0.49 & 0.87 & \textcolor{TUdark_light}{\texttt{inf}} & \textcolor{TUdark_light}{\texttt{inf}} & \textcolor{TUdark_light}{\texttt{inf}} & \textcolor{TUdark_light}{\texttt{inf}} & \textcolor{TUdark_light}{1.82\textsuperscript{\textdagger}} & 2.47 & \textcolor{TUdark_light}{1.35\textsuperscript{\textasteriskcentered}} & \textcolor{TUdark_light}{\texttt{inf}} & 2.32 & 0.46 \\
\sinvnum & \textcolor{TUdark_light}{\texttt{NaN}} & \textcolor{TUdark_light}{\texttt{NaN}} & 0.45 & 0.80 & \textcolor{TUdark_light}{\texttt{inf}} & \textcolor{TUdark_light}{\texttt{inf}} & \textcolor{TUdark_light}{\texttt{inf}} & \textcolor{TUdark_light}{\texttt{inf}} & \textcolor{TUdark_light}{2.55\textsuperscript{\textdagger}} & 2.21 & \textcolor{TUdark_light}{1.50\textsuperscript{\textasteriskcentered}} & \textcolor{TUdark_light}{\texttt{inf}} & 1.82 & 0.44 \\
\bottomrule
\end{tabularx}}

     \end{subtable} 
\end{table}

\textbf{\resnet \wl.}
The \resnet \wl is notable, since only one submission, \generalizedadam, could \emph{reliably} train it to the target performance within the given budget.
This may be surprising, as the \resnet target threshold was derived using the same procedure as all other \wls.
The \hp search space used for target-setting \citep[see][]{Dahl2023AlgoPerf} may have been more suitable for this \wl given its well-studied nature.
Although only one submission achieved a finite \resnet \wl score (recall a finite \wl score requires at least $3$ out of the $5$ repetition studies to train successfully), several others met the target in at least one study.
\nadamp, \shampoosubshort, and the \baseline hit the target at least once but were not consistent, with the remaining studies falling just short (\cref{fig:resnet}).
These results support using the median across multiple studies to determine a submission's \wl score, ensuring robustness against stochasticity.
Across all \wls, including \heldout ones, at least one submission reliably reached the target, demonstrating that none of the \wls were impossible to handle, rather it was difficult for a submission to do well on all of them simultaneously.

\textbf{Speed-up comparisons.}
In addition to benchmark scores, we compute raw speed-ups between submissions.
For relative training speed-ups, we focus on the fixed base \wls and impute infinite workload scores with the ruleset's runtime budget, \ie assuming the submission would have achieved the target just after the (artificial) cut-off.
This is a best-case assumption for submissions that do not train successfully (and thus a worst-case assumption for speed-ups over those submissions), which may be reasonable in some cases (\eg \cref{fig:resnet}) but can significantly underestimate actual training times in others.
For speed-up calculations, we do not enforce the held-out \wl rules that would invalidate \wl \runtimes when the corresponding \heldout \wl was not successfully trained.
\Cref{tab:speedups} lists the average speed-ups over the baseline across all eight base \wls.
\shampoosub provides significantly accelerated training than the external tuning \baseline, with an average speed-up of $\approx\! 28\%$.
It was also $\approx\! 2\%$ and $19\%$ faster compared to the second- and third-place submissions, respectively.
In the self-tuning ruleset, \sfadam provides $\approx\! 8\%$ faster hyperparameter-free training than the self-tuning baseline.
While speed-up metrics offer a more intuitive measure, they are less meaningful overall, as the penalty for missing a target is relatively weak. For instance, \caspr achieved a finite workload score on only four out of eight \wls, but ranked third in average speed-up.

\subsection{Comparison across \hp tuning rulesets}
\label{sec:results_comparisons}
To compare submissions across rulesets, we evaluated the first-place self-tuning submission as an external tuning submission.
This quantifies the performance gap between rulesets and estimates the expected slowdown from the lack of parallel \hp tuning.
The self-tuning winner, the (self-tuning) \sfadam submission, would have scored a $0.4804$ under the external tuning ruleset, ranking hypothetically eighth, without affecting the scores of other external tuning submissions.
Although there is a notable gap compared to its external tuning counterpart, it is surprisingly competitive in this ruleset, given its lack of parallel tuning.
Across the seven base \wls both submissions trained successfully,
\shampoosub, the winner of the external tuning ruleset, is $\approx\! 24\%$ faster than (self-tuning) \sfadam.
However, (self-tuning) \sfadam is $\approx\! 10\%$ faster than the (external tuning) \baseline across the seven base \wls both trained successfully.
In other words, (self-tuning) \sfadam reduced training time by $10\%$ compared to the baseline, using only a single machine instead of five in parallel exploring different \hp configurations.
Although there is still a meaningful gap between the rulesets, the encouraging results of (self-tuning) \sfadam suggest that even without any free \wl-specific tuning, it might be possible to create training algorithms that are surprisingly effective.

\subsection{Rule change counterfactuals}
\label{sec:results_rules}

In order to better understand the effects of the specific \algoperf benchmark rules, we can consider how the results would have changed with different rules.

\textbf{Different ways of determining whether a submission receives a finite runtime score.}
There are a variety of mechanisms in the rules whereby a submission can fail to receive a finite runtime score on a given \wl, and they triggered relatively frequently in the competition.
In the external tuning ruleset, out of \num{154} \wl-submission combinations (including both fixed base and \heldout \wls), \num{33} \iclrfinaltext{($\approx\! 21\%$)} were classified as \texttt{inf}s due to a submission not reaching the target (\num{31} out of \num{84}\iclrfinaltext{ or $\approx\! 37\%$} for the self-tuning ruleset), while \num{5}\iclrfinaltext{ ($\approx\! 3\%$) } and \num{4}\iclrfinaltext{ ($\approx\! 5\%$)}, respectively, were classified as \texttt{NaN}s (error before first evaluation).
Additionally, \num{8} held-out \wl scores\iclrfinaltext{ ($\approx\! 12\%$)} in the external tuning ruleset were ignored due to failure to reach the target on the base \wl, while the reverse occurred for \num{7} base \wl scores\iclrfinaltext{ ($\approx\! 8\%$)}.
In the self-tuning ruleset, \num{3} held-out scores\iclrfinaltext{ ($\approx\! 8\%$)} and \num{2} base scores\iclrfinaltext{ ($\approx\! 4\%$)} were excluded for the same reasons.
Finally, \num{11} scores\iclrfinaltext{ ($\approx\! 7\%$)} in the external and \num{4} scores\iclrfinaltext{ ($\approx\! 5\%$)} in the self-tuning ruleset exceeded the four-times-the-fastest-valid-runtime threshold and were thus classified as infinite.
This is most notable for \ogbg, where the fast training times of \caspr invalidated a significant number of workload scores.
Ultimately, requiring training algorithms to be robust to workload variations through the \heldout \wls \emph{did} provide a non-trivial constraint, though the \heldout \wl targets were not so stringent the effect was that large. Only considering submissions to train a \wl successfully when no other submission could train more than four times faster, primarily affected \wls where the overall \runtime budget was the most generous (in hindsight). This rule might become redundant if \runtime budgets for future iterations are tightened based on the top-performing submissions.
In \cref{fig:scores_max_tau} (\cref{app:additional_results}), we investigate the sensitivity of benchmark scores and rankings to changes in $\tau_{\text{max}}$, the upper limit of the performance profile.

\textbf{Scoring on a subset of \wls.}
How would the competition results change if we considered only a subset of the benchmark's \wls?
\Cref{fig:perf_profiles_ignore_heldouts} shows the performance profiles and benchmark scores when ignoring all \heldout \wls.
With the exception of \caspr and \amos, which switch positions, \heldout \wls have little impact on the leaderboard.
However, quantifying the full effect of \heldout \wls is challenging.
Their presence may have discouraged submissions overfitting to the fixed \wls.
While \heldout \wls offer some value, their costs likely outweigh their benefits (\cref{sec:lessons}).
These costs include the additional compute and, perhaps more importantly, the significant human effort involved in designing and implementing them.

\citet{Dahl2023AlgoPerf} envisioned \algoperf's qualification set as a cheaper way to evaluate \tas, allowing for better allocation of compute resources to the more promising submissions.
For the competition, we had enough compute to fully score all submitted \tas and did not use the qualification set (for a description of all changes between the benchmark and competition see \cref{sec:methods_mods}).
\Cref{fig:perf_profiles_qualification_set} shows the submissions' score when using only the qualification set (which also excludes \heldout \wls).
While the qualification set helps to identify under-performing submissions, rankings can change substantially compared to the full set.
We can also assess how scores shift when removing individual \wls.
\cref{tab:leaderboard_ignoring_wls} presents the benchmark scores and rankings when each \wl is excluded.
The rankings remain largely stable, with the winning submissions unchanged in all but one case, indicating that the competition results are robust to the precise selection of \wls, as long as there is a large and diverse enough set.

\section{Engineering challenges}
\label{sec:engineering}

Scoring submissions based on wall-clock time, especially in a multi-framework setting, introduces non-trivial engineering challenges. Requiring competing submissions to make use of shared \wl implementations and timing code is an essential requirement for meaningful timing measurements, but also introduces a host of issues. Even for standard algorithms, seemingly minor implementation details can have a dramatic effect on time-per-step or how quickly the loss decreases per step. As much as possible, potential confounding factors for training algorithm comparisons should be removed, while still allowing realistic participation. Perhaps the biggest point of tension between controlling measurements and remaining relevant to current practice is due to supporting multiple frameworks. Submissions in \jax \& \pytorch have to be compared on a level playing field, despite framework-specific differences in optimization and execution paradigms.
To ensure as fair a comparison as possible between algorithms implemented in \jax \& \pytorch, the workload implementations in each framework should be functionally equivalent, high-quality, and realistically performant:

\textbf{Functional equivalence}: \wl implementations should perform mathematically identical computations across frameworks and use identical implementation strategies. Given the same input (batch, model parameters, random initialization), the outputs of the forward and backward passes should be the same (within numerical precision) regardless of whether the \wl is implemented in \jax or \pytorch. All parts of the computation not controlled by the submission code (everything from forward \& backward passes to weight initialization to timing \& logging) should be semantically identical and also be parallelized in the same way on the same devices. This guarantees that any observed performance differences are truly due to algorithmic improvements within the submissions, and not variations in the underlying computations.

\textbf{Performant in both time and memory}: \wl implementations shouldn't saddle training algorithm submissions with time or memory overheads that skilled engineers would easily avoid outside the context of the competition. Although achieving identical execution times (and memory usage patterns) across frameworks when run on identical competition hardware is not required in principle (and is nearly impossible), in practice since both \jax \& \pytorch are mature frameworks and the competition \wls are standard deep learning benchmark problems, any large discrepancy between time or space efficiency across \wl implementations in different frameworks is cause for concern. Hypothetically, different frameworks can make different techniques easier or harder to express or make achieving certain memory or runtime requirements more or less difficult. However, only when the \wl implementations in \emph{both} frameworks are as efficient as is realistically possible can one begin to ask how the constraints of the frameworks themselves are playing a role. Implementations should realistically capture how the frameworks are used by practitioners while also being as efficient as possible.

Addressing these challenges led to the identification and subsequent implementation of various improvements and best practices. Our experience should be useful for researchers attempting to make similar reproducible measurements with as few confounding factors as possible.

\subsection{Functional equivalence of \jax \& \pytorch implementations}
\label{subsec:functional_equivalence}

\textbf{Framework-specific defaults.} \jax \& \pytorch provide basic primitives for implementing various components in neural networks. Creating functionally equivalent \wls requires considerable care around framework specific features and defaults. For example, \jax's default \texttt{Gelu} activation function calculates the cumulative density function of the Gaussian with a \texttt{Tanh} approximation, whereas in \pytorch the exact computation is used. Similarly, for layer normalization, the default values for $\varepsilon$ differ between the frameworks. Even more insidious, the weights of linear layers in \jax are initialized with \texttt{lecun\_normal}, while similar weights in \pytorch are initialized with \texttt{kaiming\_uniform}. Moreover, \pytorch does not supply a \texttt{lecun\_normal} initialization, necessitating a manual implementation.

\textbf{Data pre-processing and augmentation.} Differences in data pre-processing and augmentation strategies can result in data pipelines that are not comparable. To mitigate these types of differences, the same \tensorflow \citep{tensorflow2015} dataset pipelines were used across frameworks where possible (\criteo, \fastmri, \librispeech, \ogbg, and \wmt). However, the \imagenet \vit and \resnet \wls in \pytorch use a custom implementation to match the \tensorflow random augmentation strategy in the \jax workload.

\subsection{Performance of \jax \& \pytorch implementations}
\label{subsec:performance}

\textbf{Data pipelines.} Differences in the data parallelism implementations may also lead to additional overhead for \pytorch workloads. To match \jax's data parallel paradigm, the \pytorch workloads are implemented with \texttt{DistributedDataParallel} (\texttt{DDP}). \texttt{DDP} requires one \python process per device, which in a naive implementation would lead to replication of \tensorflow data pipeline tasks in each process and  potentially use too much RAM and an undesirable number of threads. To mitigate this issue, \tensorflow data pipeline operations are only run in one \python process and batches are broadcasted to the remaining processes. This choice results in a small additional communication overhead for each batch in the \pytorch workloads.

\textbf{Custom CUDA kernels.} The wall-clock time of two functionally equivalent computations can significantly differ due to differences in the underlying GPU implementations. One example for this is the \texttt{torch.lstm} implementation. In \pytorch, the LSTM implementation uses a custom CUDA kernel that results in a \num{2}$\times$ smaller wall-clock time on the \librispeech \deepspeech workload compared to the \jax version. This uncovered the opportunity to generate a similar CUDA kernel in the \jax LSTM layer implementation. Using the updated faster \jax LSTM layer implementation significantly sped up the end-to-end wall clock times for the \deepspeech \wl. \iclrfinalfootnote{\url{https://github.com/google/jax/pull/13319}} 

\textbf{Adopting \pytorch 2.0 features} Achieving realistically performant \wl implementations requires adopting novel features and best practices in both frameworks. Without these, the performance gaps between \pytorch \& \jax could be as large as 60\%.
Over the course of the development of the competition, the performance of \pytorch workload implementations significantly improved as a result of migrating from \pytorch to \pytorch 2.0 (see \cref{sec:timing_frameworks}).

\pytorch 2.0 \citep{10.1145/3620665.3640366}, released in Dec 2022, introduced two major extensions that represent a major departure from \pytorch's \citep{Paszke2019} original eager programming model, where every line of code would dispatch to a CUDA kernel. Specifically, \pytorch 2.0 introduced \texttt{TorchDynamo}, a Python level JIT compiler that enables graph compilation of \pytorch programs, and \texttt{TorchInductor}, a compiler backend which translates \pytorch programs into Triton \citep{10.1145/3315508.3329973} kernels for GPU and C++ for CPUs. 
The main adjustments that had to be made to adopt to \pytorch 2.0 and close the speed gap to \jax were related to \texttt{torch.compile}:

\begin{itemize}[itemsep=0pt,leftmargin=1em]
    \item \textbf{Avoiding graph breaks.} When \texttt{TorchDynamo} encounters unsupported functionality, it creates a graph break, splitting the model. To maximize performance, graph breaks should be prevented and models compiled with \texttt{torch.compile(model, fullgraph=True)}.
    \item \textbf{Compiling the loss function, not just the model.} Most of the tutorials surrounding \texttt{torch.compile} implied that it is only to be applied to the model by running \texttt{torch.compile(model)}. However, compiling loss functions had a dramatic impact on performance for several workloads.\iclrfinalfootnote{\url{https://github.com/mlcommons/algorithmic-efficiency/pull/597}} 
    \item \textbf{Overhead reduction mode.} For models where overhead reduction is crucial it is recommended to use CUDA graphs. This can be enabled via \texttt{torch.compile(model, mode="reduce-overhead")} for the cost of a small memory overhead.
    \item \textbf{Combining \texttt{DDP} and \texttt{torch.compile}}: The compiler also needs to compose with the other subsystems, e.g., there are subtle differences between \texttt{ddp(torch.compile(model))} and \texttt{torch.compile(ddp(model))}. With the latter, the compiler will also trace collectives which may result in further performance optimizations. \end{itemize}

\textbf{Memory allocator settings} Older CUDA versions are more likely to cause OOM errors in \pytorch. Setting \texttt{torch.cuda.memory.\_set\_allocator\_settings('expandable\_segments:True')} can fix OOMs caused by memory fragmentation.\iclrfinalfootnote{\url{https://github.com/mlcommons/algorithmic-efficiency/issues/497}}

Making adjustments to use modern features and adopting best practices for \pytorch \& \jax helped achieve realistically performant \wl implementations in both frameworks and significantly reduced the time gap in between the \jax \& \pytorch \wl implementations. While performance differences still exist, they are mostly within tolerance (12\%) and are balanced such that neither framework has an overwhelming advantage. Measurements of the performances gaps between \pytorch \& \jax \wls and improvements can be found in \cref{sec:timing_frameworks}.

\section{Lessons learned}
\label{sec:lessons}

The improved {\ta}s developed for the competition delivered significant speedups in \nn training.
The winning submissions achieved $28\%$ and $8\%$ faster model training compared to their respective baselines.
The winning submission in the external tuning ruleset, based on \textsc{Distributed Shampoo}, demonstrates that non-diagonal preconditioning methods can improve runtime over currently popular methods like \adam.
And yet, \textbf{despite these significant strides made in accelerating \nn training, there is ample room left for algorithmic improvement}.
No single submission dominated across all \wls; instead, five different submissions achieved the best performance on at least one of the eight base \wls.

\textbf{These algorithmic advances can only be realized reliably by careful benchmarking and engineering efforts ensuring fair and meaningful comparisons}.
Comparisons on multiple \dl \wls are required to isolate a robust signal of a submission's performance.
Seemingly intuitive aggregate metrics, like average speedup, fail to fully capture pertinent aspects of a \ta's practical usefulness.
Precise engineering work is required to ensure that \tas are compared fairly. 
In \Cref{sec:engineering}, we identified implementation details that dramatically affect the runtime across virtually all \tas, and numerous potentially confounding factors that must be accounted for in algorithmic comparisons, in particular across \dl frameworks.

The competition also underscores the inherent link between \hp tuning and \ta performance.
\textbf{To meaningfully compare \tas, measurements must properly account for \wl-specific \hp tuning and \tas must be fully-specified without leaving free parameters for the user to set.} A complete \ta specification includes a formal specification of a \ta's \hp defaults and/or search space, and should include regularization choices.
The varying performance of many submissions across \wls suggests that \hp tuning remains a significant challenge; the top-scoring submissions mostly distinguished themselves by their reliable training across a large variety of \wls.
Despite promising advances in \hp-free algorithms (as seen in the self-tuning ruleset), fully-automatic \nn training remains a serious challenge.
In light of our results, future publications of \tas should include clear \hp (search space) recommendations and be paired with a recommended tuning protocol, ideally one that is sensitive to a user-supplied tuning budget. Although a radical change from the current practice of published \tas that aren't runnable without setting various \hps, publishing families of update rules and abdicating responsibility for tuning entirely to the user only adds to the community's confusion on what to actually use.

\subsection{Methodological lessons}
\label{sec:lessons_methods}

In our competition, the \algoperfTA benchmark has proven to be quite effective in differentiating algorithms and measuring progress in \nn training, but it results in quite an involved experimental protocol with substantial costs. Most of its features, e.g. (integrated) performance profiles, were useful in order to generate nuanced and robust insights. However, based on our experience, we propose the following modifications to the \algoperf benchmark going forward:
(1) \textbf{Removing \heldout \wls.} Eliminating the \heldout \wls would drastically simplify the evaluation process and reduce the benchmark's runtime substantially.
Though \heldout \wls have the potential to deter overfitting to the base \wls, they require substantial computational, logistical (they can't be generated until submissions are frozen), and engineering effort.
Replacing the six \heldout \wls with one or two additional base \wls would provide similar benefits, while reducing runtime and allowing additional, practically-relevant \wls to be included, such as autoregressive language models or diffusion models.
(2) \textbf{Reducing the runtime budgets to reduce costs, especially for the self-tuning ruleset.} 
The submissions have demonstrated that, for many \wls, significantly less training time is needed.
Matching the self-tuning budget to the current external tuning budget, and potentially reducing the external tuning budget further, would maintain meaningful comparisons while lowering overall costs.
(3) \textbf{Reducing the number of studies from 5 to 3} would cuts compute costs by an additional 40\%.
These repetitions with different random seeds ensure robust insights rather than random noise due to the stochastic training process.
However, our results indicate fewer studies are sufficient for statistical fidelity.
Additionally, a modernized hardware setup with a better cost-to-performance ratio could further reduce costs.

\section{Conclusion}
\label{sec:conclusion}

An inescapable limitation of empirical comparisons of (training) algorithms \citep[see also][]{Choi2019,Sivaprasad2020,Schmidt2021} is that one can only directly measure the behavior of specific implementations, not the abstract algorithmic ideas they express. The \algoperfTA competition embraces this constraint by evaluating submissions that fully specify everything necessary to apply them to any generic learning problem, including all necessary \wl-specific \hp tuning. The competition conditions provide a realistic simulation of applying a generic \ta to a new problem, with some important caveats. First and foremost, the maximum runtime budgets and evaluation metric targets for each workload provide a lot of information about what training horizon is appropriate and what validation error is achievable, unlike in a novel learning task where such information isn't available. Second, by scoring on time to reach particular \emph{validation} error goals and not using results on separate per-\wl test sets, the competition conditions do not capture the challenges of train/test skew or distribution shift.
Additionally, while this competition reveals \emph{how well} training algorithms perform, deeper analysis, such as \citet{Kunstner2024}, is needed to explain \emph{why}.

Nevertheless, despite its limitations, the \algoperfTA competition has produced insights about the current \ta landscape. In particular, the competition results have validated two exciting directions for improvements: non-diagonal preconditioning (\shampoosub) and \hp reduction strategies (\sfadam). These insights are only possible because they are the result of an open and \emph{competitive} process. This is in contrast to existing works that, due to the lack of a suitable existing benchmark, have each been forced to introduce their own \ta evaluation protocols. As a result, much of existing research on training algorithms has focused on isolating and improving individual ideas and components, but hasn't always produced complete and usable \tas that incorporate viable \hp tuning protocols, as would be necessary to perform well in competition.

Our end goal is not the specific insights generated by the inaugural round of comparisons, as interesting as they might be. 
Ideally, competitions like \algoperf will affect \tas research in two main ways. First, they should provide a sieve to filter the most useful ideas out of the literature by separating truly practical methods from interesting ideas that aren't yet useful. Second, they should provide a valuable signal to guide the design process for new \tas.

\newpage

\subsubsection*{Acknowledgments}
FS was supported by funds from the Cyber Valley Research Fund.
PH and FS gratefully acknowledge co-funding by the European Union (ERC, ANUBIS, 101123955), and by the DFG through Project HE 7114/5-1 in SPP2298/1;
PH is a member of the Machine Learning Cluster of Excellence, funded by the DFG under Germany's Excellence Strategy – EXC number 2064/1 – Project number 390727645; 
PH and FS also acknowledge the German Federal Ministry of Education and Research (BMBF) through the Tübingen AI Center (FKZ:01IS18039A); and funds from the Ministry of Science, Research and Arts of the State of Baden-Württemberg. 

\bibliography{bibliography}
\bibliographystyle{iclr2025_conference}

\newpage  \appendix
\section{Appendix}

\subsection{\algoperf details}

In this section, we provide additional details about the \algoperfTA benchmark \citep{Dahl2023AlgoPerf} and our competition based on it.
\Cref{tab:workloads,tab:held_out_workloads} summarize the fixed base and \heldout \wls of the \algoperf benchmark as they are used in the competition.

\begin{table}[!htp]
    \caption{\textbf{Summary of \emph{fixed} base \wls in the \algoperf benchmark.} Losses include cross-entropy (CE), mean absolute error (L1), and Connectionist Temporal Classification loss (CTC). Additional evaluation metrics are structural similarity index measure (SSIM), (word) error rate (ER \& WER), mean average precision (mAP), and bilingual evaluation understudy score (BLEU). Note: Some \wls have minor changes (see \cref{sec:methods_mods}) to the \runtime budgets and validation targets compared to the \algoperf benchmark publication \citep{Dahl2023AlgoPerf}. The \runtime budget is that of the external tuning ruleset, the self-tuning ruleset allows $3\times$ longer training.}
	\label{tab:workloads}
	\centering
	\small
	\begin{tabularx}{0.98\textwidth}{@{}XllllS[table-format=2.6]S[table-format=6.0]@{}}
	\toprule
                              &              &              &     &      & {Validation} & {\Runtime} \\ 
\textbf{Task} & \textbf{Dataset} & \textbf{Model} & \textbf{Loss} & \textbf{Metric} & \textbf{Target} & \textbf{Budget} \\ \midrule
Clickthrough rate \newline prediction  & \criteo      & \dlrmsmall   & CE  & CE   & 0.123735   & 7703    \\ \addlinespace
MRI reconstruction            & \fastmri     & \unet        & L1  & SSIM & 0.7344     & 8859    \\ \addlinespace
Image                         & \imagenet    & \resnetfifty & CE  & ER   & 0.22569    & 63008   \\
classification                &              & \vit         & CE  & ER   & 0.22691    & 77520   \\ \addlinespace
Speech                        & \librispeech & \conformer   & CTC & WER  & 0.085884   & 61068  \\
recognition                   &              & \deepspeech  & CTC & WER  & 0.119936   & 55506   \\ \addlinespace
Molecular property \newline prediction & \ogbg        & \gnn         & CE  & mAP  & 0.28098    & 18477   \\ \addlinespace
Translation                   & \wmt         & \transformer & CE  & BLEU & 30.8491    & 48151   \\ \bottomrule
\end{tabularx}
 \end{table}

\begin{table}[!htp]
    \caption{\textbf{Summary of \emph{held-out} \wls sampled for this iteration of the \algoperf competition.} One held-out \wl was sampled from the random \wl variants (see \citep[Table 11]{Dahl2023AlgoPerf}) for each \dataset used in the fixed \wls. Note, for the \imagenet and \librispeech \datasets only a single held-out \wl was sampled, according to the modified competition rules (see \cref{sec:methods_mods}). The loss function and performance metrics are identical to the corresponding fixed base \wl (\cref{tab:workloads}).}
	\label{tab:held_out_workloads}
	\centering
	\small
	
\begin{tabularx}{0.98\textwidth}{@{}XllS[table-format=2.6]S[table-format=6.0]@{}}
	\toprule
                              &                  &                             & {Validation}      & {Runtime}         \\ 
\textbf{Task}                 & \textbf{Dataset} & \textbf{Variant}            & \textbf{Target}   & \textbf{Budget} \\ \midrule
Clickthrough rate prediction  & \criteo          & \textsc{Embed Init Scale}     & 0.129657        & 7703              \\ \addlinespace
MRI reconstruction            & \fastmri         & \textsc{TanH}                 & 0.717840          & 8859              \\ \addlinespace
Image classification          & \imagenet        & \textsc{ResNet BN Init Scale} & 0.23474         & 63008             \\ \addlinespace
Speech recognition            & \librispeech     & \textsc{Conformer LayerNorm}  & 0.09731         & 61068             \\ \addlinespace
Molecular property prediction & \ogbg            & \textsc{Altered Layers}       & 0.269446         & 18477             \\ \addlinespace
Translation                   & \wmt             & \textsc{GLU \& TanH}          & 29.5779         & 48151             \\ \bottomrule
\end{tabularx}
 \end{table}

\subsection{Modifications to the \algoperf benchmark for this competition}
\label{sec:methods_mods}

For the purposes of this competition, we made several modifications to the benchmark from how it was initially proposed by \citet{Dahl2023AlgoPerf}, mainly to reduce the competition's overall compute costs.
All modifications were done before the competition's submission deadline.
Most notably, the number of tuning trials in the external tuning ruleset was reduced from $20$ to $5$.
This change significantly lowers the computational cost of the competition while also mitigating the risk of submissions overfitting to the benchmark's \wl selection.
To further reduce computational costs, we also decreased the \runtime budgets for both \librispeech \wls significantly---from $101,780$ to $61,068$ seconds and from $92,509$ to $55,506$ seconds---with only a slight impact on target performance (the WER changed from $0.078477$ to $0.085884$ and from $0.1162$ to $0.119936$).
Additionally, small bug fixes led to slight changes in two other \wl targets: the target loss on \criteo increased from $0.123649$ to $0.123735$, and the target SSIM on \fastmri decreased from $0.7344$ to $0.723653$.
Furthermore, only a single held-out \wl was sampled for each \emph{\dataset}, not each \wl.
This means that for both the \imagenet and \librispeech \datasets, only one, instead of two, held-out \wls were sampled from the combined set of \wl variants, further reducing the competition's overall \runtime significantly.

Finally, we decided to use only the validation targets, excluding the test set targets.
Including both validation and test set targets, as proposed by \citet{Dahl2023AlgoPerf}, complicates the evaluation process by involving the practitioner's task of selecting \hps for an unknown test set based on validation performance.
We chose to omit this aspect of the training process in the current iteration of the benchmark.

\subsection{Computational costs}
\label{app:costs}
The \algoperf benchmark requires a substantial number of training runs and thus significant computational resources to yield meaningful results.
To evaluate a single submission, it needs to be evaluated on eight fixed---and six held-out---\wls.
For each \wl, five repetition studies are performed, and in the external tuning ruleset, each study comprises five tuning trials.
In this iteration of the competition, we conducted \num{3850} runs in the external tuning ruleset (due to eleven \tas evaluated on $14$ \wls, with $5\!\cdot\!5$ trials) and \num{420} runs in the self-tuning ruleset (six \tas evaluated five times on $14$ \wls, with a runtime budget $3\!\times$ larger).
The actual computational cost of these \num{4000}+ training runs varies significantly based on the performance of the submission.
Training can terminate early if the validation (and test targets)\footnote{Although we decided not to use test targets for the purpose of our competition, we wanted to log when submissions reached the test targets for potential future analysis.} are met or if errors occur, such as out-of-memory (OOM) issues.
The runtime budget in \cref{tab:workloads} reflects only the compute time spent by the submissions themselves, excluding time for free evaluations or other operations outside the submission functions.
It is thus not an upper bound on the total computational cost.
On average, we required approximately \SI{1847}{\hour} to fully run a self-tuning submission and \SI{3469}{\hour} per external tuning submission, amounting to a total of roughly \SI{49240}{\hour} on the competition hardware (\benchmarkinghardware).
This is considerably less than what is reported in \citet[Table 15]{Dahl2023AlgoPerf}, primarily due to fewer tuning trials and a reduced runtime budget for certain \wls (see \cref{sec:methods_mods}), as well as early termination of some submissions.
In theory, the number of required runs could be reduced, for example, by skipping a \heldout \wl if a submission fails the corresponding fixed \wl.
However, this would limit the ability to parallelize different runs and reduce opportunities for further analysis beyond calculating benchmark scores.

\subsection{Submission details}

In this section, we provide details on the submissions received in this iteration of the competition.
\Cref{tab:submissions_et} lists the submissions to the external tuning ruleset, while \cref{tab:submissions_st} details submissions received for the self-tuning ruleset. Notably, no submission in this iteration significantly modified the \texttt{data\_selection} function, indicating a potential area for future exploration.

\begin{table}[!htp]
    \caption{\textbf{External tuning submissions.} Details of all submissions to the external tuning ruleset.}
	\label{tab:submissions_et}
	\centering
	\small
	{\renewcommand{\arraystretch}{1.3}
\setlength{\tabcolsep}{4pt}
\hyphenpenalty=10000\exhyphenpenalty=10000
\begin{tabularx}{0.98\textwidth}{@{}p{1.5cm}>{\raggedright}p{2.5cm}>{\raggedright}p{1.5cm}p{1.5cm}X@{}}
\toprule
\textbf{Submission} & \textbf{Authors} & \textbf{Institutions} & \textbf{Framework} & \textbf{Description} \\ \midrule
\makecell{\textsc{PyTorch}\\ \textsc{Distr.}\\ \textsc{Submission}} & Hao-Jun Shi, Tsung-Hsien Lee, Anna Cai, Shintaro Iwasaki, Wenyin Fu, Yuchen Hao, Mike Rabbat & Meta Platforms  & \pytorch & Based on the Distributed Shampoo algorithm of \citet{anil2020shampoo} with an implementation tailored to leverage PyTorch performance optimizations. See~\cite{shi2023distributeddataparallelpytorchimplementation} for details. The submission uses a list of five hyperparameter settings. \\
\makecell{\textsc{Schedule} \\ \textsc{Free}\\ \textsc{AdamW}} & Alice Yang, Aaron Defazio, Konstantin Mishchenko & Meta AI, Samsung AI  & \pytorch & A externally tuned version of \sfadam \citep{defazio2024road} with a list of five \hp configurations. \\
\makecell{\textsc{General-}\\ \textsc{ized}\\ \textsc{Adam}} & George Dahl, Sourabh Medapati, Zack Nado, Rohan Anil, Shankar Krishnan, Naman Agarwal, Priya Kasimbeg, Vlad Feinberg & Google  & \jax & Submission with an \adam-style update rule, tuning over the use of Nesterov acceleration and preconditioning. Essentially tuning over \adamw \citep{Kingma2015}, \nadamw, and \sgd \citep{Robbins1951} with or without momentum.\\
\cycliclr & Niccolò Ajroldi, Antonio Orvieto, Jonas Geiping & MPI-IS, ELLIS Institute Tübingen  & \pytorch & Revisits the work of \citet{Loshchilov2017} and \citet{smith2017cyclical}, coupling \nadamw \citep{Dozat2016,Loshchilov2019} with a cyclic learning rate scheduler. Each cycle involves a linear warmup phase for the LR, followed by cosine annealing. \\
\nadamp & George Dahl, Sourabh Medapati, Zack Nado, Rohan Anil, Shankar Krishnan, Naman Agarwal, Priya Kasimbeg, Vlad Feinberg & Google  & \jax & Uses \nadamw with an extra tunable hyperparameter $p$ enabling $p$th root of denominator inside \nadamw update rule instead of the default of $2$. \\
\rowcolor{TUgray_light}\baseline & &  & \jax &  Baseline using \nadamw \citep{Dozat2016,Loshchilov2019} and a linear learning rate warmup followed by a cosine decay \citep{Dahl2023AlgoPerf}. \\
\amos & Ran Tian & Google & \jax & Submission based on the \amos optimizer \citep{Tian2022} with a list of five \hp settings.\\
\makecell{\textsc{CASPR}\\ \textsc{Adaptive}} & Sai Surya Duvvuri, Inderjit S. Dhillon, Cho-Jui Hsieh & UT Austin, UCLA, Google  & \jax &  A submission based on \citep{Duvvuri2024} with a list of five \hp configurations.\\
\lawaq & Niccolò Ajroldi, Antonio Orvieto, Jonas Geiping & MPI-IS, ELLIS Institute Tübingen  & \pytorch & Employs Latest Weight Averaging \citep{izmailov2018swa,kaddour2022lawa} on top of \nadamw \citep{Dozat2016,Loshchilov2019}, maintaining a queue of previous model weights. The queue is periodically updated during training and passed to the competition API for evaluation. \\
\lawaema & Niccolò Ajroldi, Antonio Orvieto, Jonas Geiping & MPI-IS, ELLIS Institute Tübingen & \pytorch &  Similar to \lawaq but maintaining an exponential moving average of the model weights, which is updated periodically during training and returned to the competition API for evaluation. \\
\makecell{\textsc{Schedule}\\ \textsc{Free}\\ \textsc{Prodigy}} & Alice Yang, Aaron Defazio, Konstantin Mishchenko &  Meta AI, Samsung AI & \pytorch & Combining Schedule-free \citep{defazio2024road} with the \textsc{Prodigy} optimizer \citep{Mishchenko2024}. \\
\bottomrule
\end{tabularx}
}
 \end{table}

\begin{table}[!htp]
	\caption{\textbf{Self-tuning submissions.} Details of all submissions to the external tuning ruleset.}
	\label{tab:submissions_st}
	\centering
	\small
	{\renewcommand{\arraystretch}{1.3}
\setlength{\tabcolsep}{4pt}
\hyphenpenalty=10000\exhyphenpenalty=10000
\begin{tabularx}{0.98\textwidth}{@{}p{1.6cm}>{\raggedright}p{2.5cm}>{\raggedright}p{1.5cm}p{1.5cm}X@{}}
\toprule
\textbf{Submission} & \textbf{Authors} & \textbf{Institutions} & \textbf{Framework} & \textbf{Description} \\ \midrule
\makecell{\textsc{Schedule}\\ \textsc{Free}\\ \textsc{AdamW}} & Alice Yang, Aaron Defazio, Konstantin Mishchenko & Meta AI, Samsung AI & \pytorch & A self-tuning version of \sfadam \citep{defazio2024road} using a single \hp configuration.\\
\rowcolor{TUgray_light}\baseline & &  & \jax & Baseline using \nadamw, a linear learning rate warmup followed by a cosine decay, and a single hyperparameter point \citep{Dahl2023AlgoPerf}. \\
\makecell{\textsc{NadamW}\\ \textsc{Sequential}} & George Dahl, Sourabh Medapati, Zack Nado, Rohan Anil, Shankar Krishnan, Naman Agarwal, Priya Kasimbeg, Vlad Feinberg & Google & \jax & Uses \nadamw update rule and runs 3 fixed \hp points sequentially. The intention was for these to be the top 3 \hp points found at one third the self-tuning ruleset step budgets. \\
\textsc{Sinv6 75} & Abhinav Moudgil & Mila, Concordia University & \jax & A submission for a task-invariant learned optimizer meta-trained on small tasks. Uses $75\%$ of the number of steps as target in learned optimizer initialization.\\
\textsc{Sinv6} & Abhinav Moudgil & Mila, Concordia University & \jax & A submission for a task-invariant learned optimizer meta-trained on small tasks. \\
\textsc{AdamG} & Yijiang Pang & Michigan State University & \pytorch & A submission based on the \textsc{AdamG} optimizer \citep{Pang2024}. \\
\bottomrule
\end{tabularx}
}
 \end{table}

\subsection{Additional competition results}
\label{app:additional_results}

\begin{figure}[!htp]
	\centering
	\begin{subfigure}[b]{.9\textwidth}
		\centering
		\includegraphics[width=\textwidth]{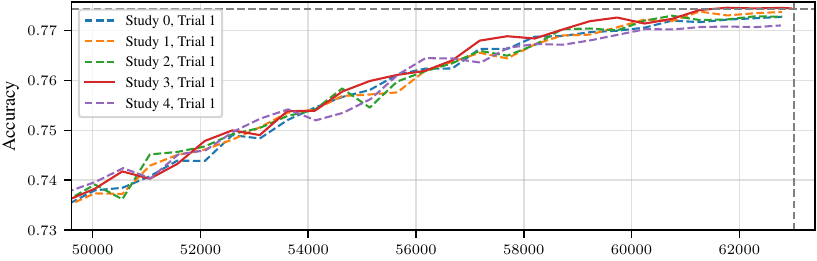}
		\caption{\baseline}
    	\label{fig:resnet_baseline}
	\end{subfigure}
	\par\bigskip \begin{subfigure}[b]{.9\textwidth}
		\centering
		\includegraphics[width=\textwidth]{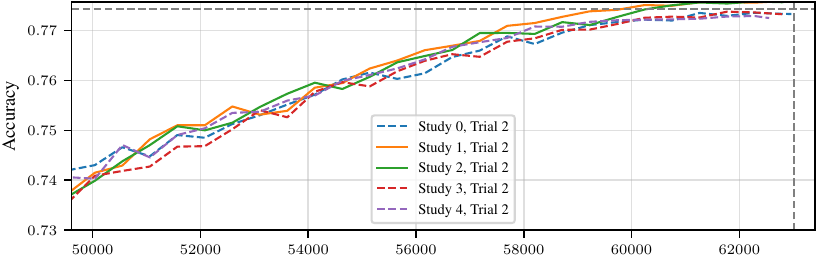}
		\caption{\nadamp}
    	\label{fig:resnet_nadamp}
	\end{subfigure}
	\par\bigskip \begin{subfigure}[b]{.9\textwidth}
		\centering
		\includegraphics[width=\textwidth]{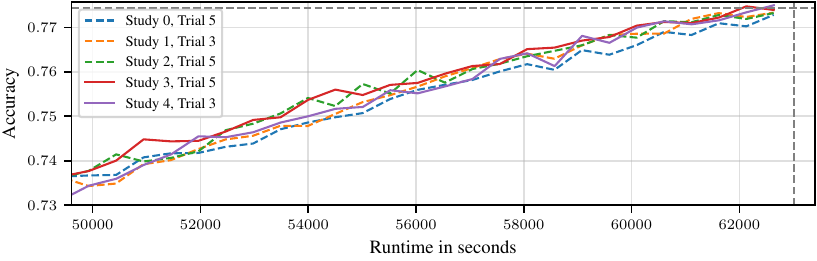}
		\caption{\shampoosub}
    	\label{fig:resnet_shampoo}
	\end{subfigure}
	\caption{\textbf{Validation accuracy vs.~runtime on the \resnet \wl.} The \baseline (\cref{fig:resnet_baseline}), \nadamp (\cref{fig:resnet_nadamp}), and \shampoosub (\cref{fig:resnet_shampoo}) all reach the validation target on the \resnet \wl for at least one study but not reliably enough to get a finite score. Shown are the best trials from each of the five studies, where ``best'' is either the fastest trial to achieve the target performance or the trial whose best performance is closest to the target. Trials that reach the target are marked with a solid line, while studies that do not reach the target are indicated with a dashed line. The gray dashed horizontal and vertical lines indicate the target performance and runtime budget respectively. Additionally, both \amos and \cycliclr came close but missed the target in all studies.}
	\label{fig:resnet}
\end{figure}

\begin{figure}[!htp]
	\centering
	\begin{subfigure}[b]{.9\textwidth}
		\centering
		\includegraphics[width=\textwidth]{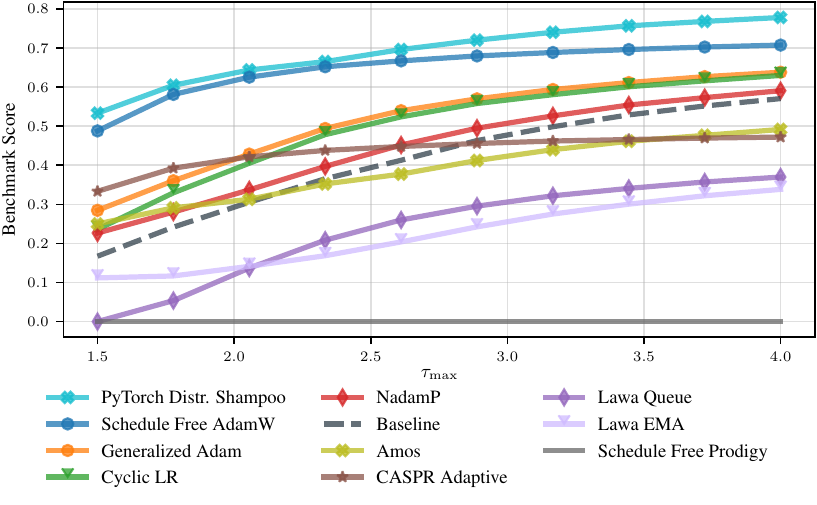}
		\caption{External tuning ruleset}
    	\label{fig:scores_max_tau_et}
	\end{subfigure}
	\par\bigskip \begin{subfigure}[b]{.9\textwidth}
		\centering
		\includegraphics[width=\textwidth]{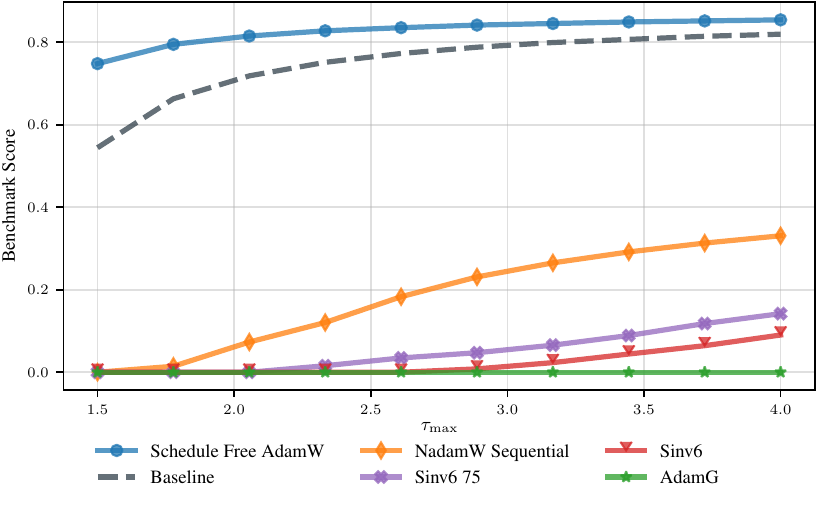}
		\caption{Self-tuning ruleset}
    	\label{fig:scores_max_tau_st}
	\end{subfigure}
	\caption{\textbf{Benchmark score as a function of $\tau_{\text{max}}$.} The upper limit of the performance profile and upper integration limit for the benchmark score, $\tau_{\text{max}}$, determines which \wl scores are treated as finite and influences the penalty for infinite scores. We observe that rankings remain stable for most submissions across different values of $\tau_{\text{max}}$.}
	\label{fig:scores_max_tau}
\end{figure}

\begin{figure}[!htp]
	\centering
	\begin{subfigure}[b]{\textwidth}
		\centering
		\includegraphics[width=\textwidth]{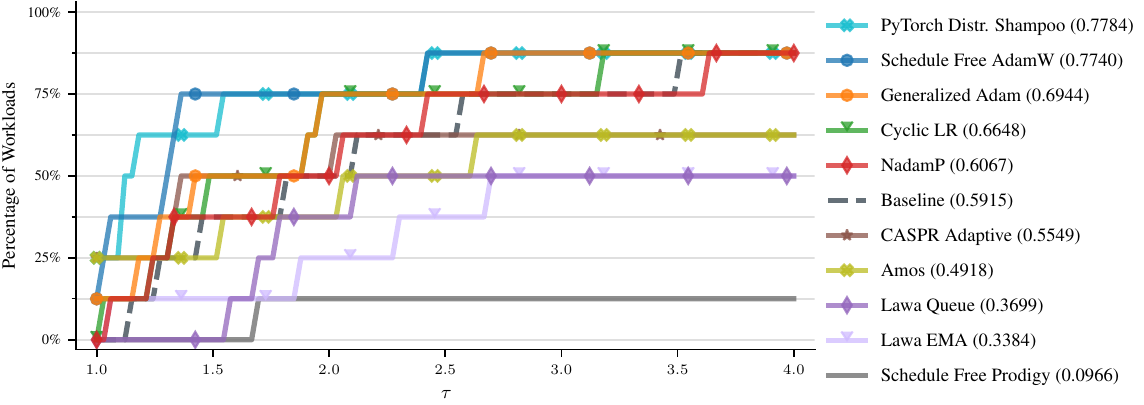}
		\caption{Performance profiles for the external tuning ruleset ignoring held-out \wls}
    	\label{fig:pp_external_tuning_ignore_heldouts}
	\end{subfigure}
	\par\bigskip \begin{subfigure}[b]{\textwidth}
		\centering
		\includegraphics[width=\textwidth]{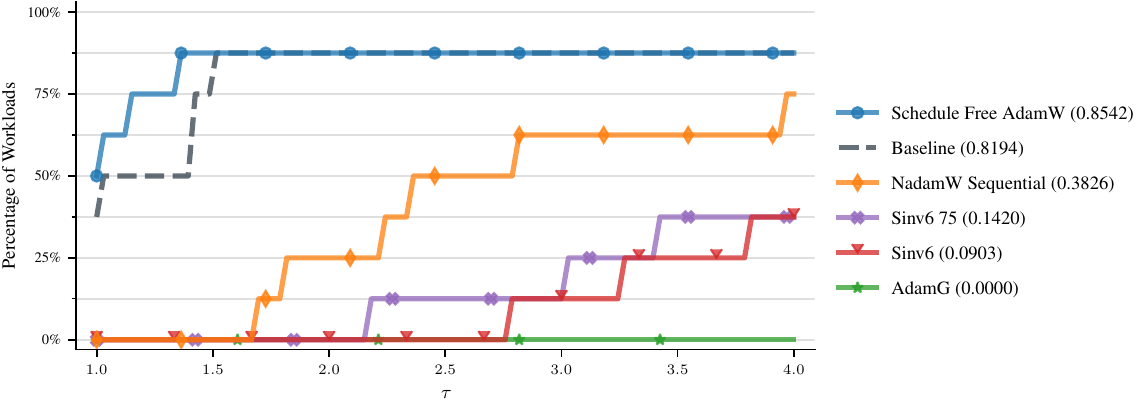}
		\caption{Performance profiles for the self-tuning ruleset ignoring held-out \wls}
    	\label{fig:pp_self_tuning_ignore_heldouts}
	\end{subfigure}
	\caption{\textbf{Performance profiles of all \algoperf submissions when ignoring held-out \wls.} Structurally the same as \cref{fig:perf_profiles} but here we ignore all benchmark rules involving the held-out \wls.}
	\label{fig:perf_profiles_ignore_heldouts}
\end{figure}

\begin{figure}[!htp]
	\centering
	\begin{subfigure}[b]{\textwidth}
		\centering
		\includegraphics[width=\textwidth]{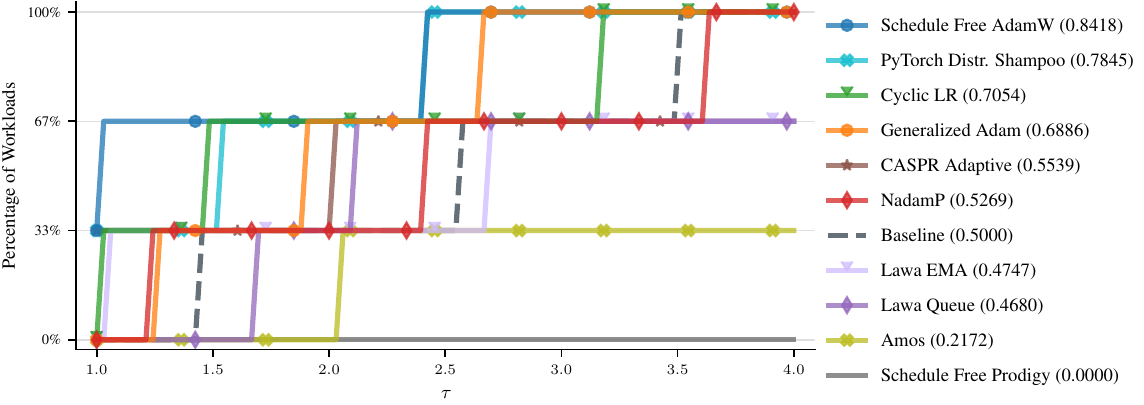}
		\caption{Performance profiles for the external tuning ruleset on the qualification \wls}
    	\label{fig:pp_external_tuning_qualification_set}
	\end{subfigure}
	\par\bigskip \begin{subfigure}[b]{\textwidth}
		\centering
		\includegraphics[width=\textwidth]{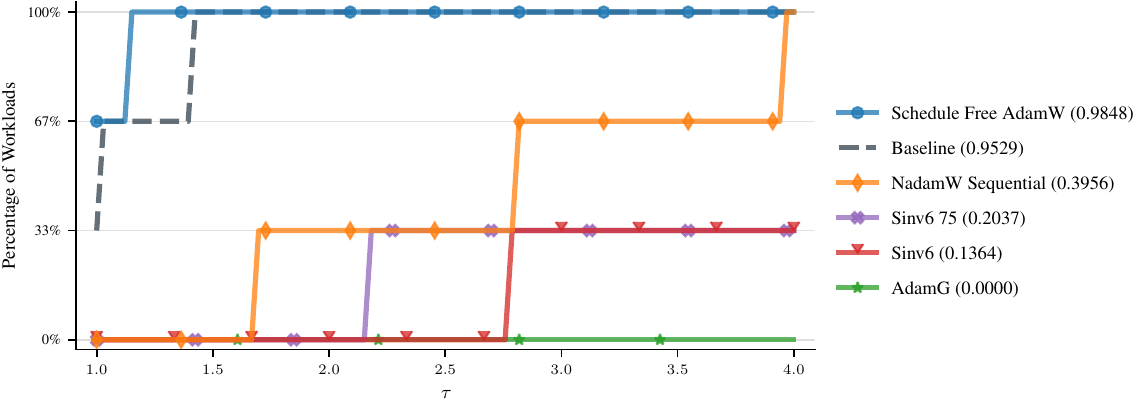}
		\caption{Performance profiles for the self-tuning ruleset on the qualification \wls}
    	\label{fig:pp_self_tuning_qualification_set}
	\end{subfigure}
	\caption{\textbf{Performance profiles of all \algoperf submissions on the qualification \wls.} Structurally the same as \cref{fig:perf_profiles} and \cref{fig:perf_profiles_ignore_heldouts} but only considering the three \wls that are part of the qualification set, \ie \criteo, \wmt, and \ogbg. In the qualification set, no held-out \wls are used.}
	\label{fig:perf_profiles_qualification_set}
\end{figure}

In this section, we provide additional analysis of the \algoperf competition results.
In \cref{fig:resnet}, we investigate submissions could achieve the target performance on the \resnet \wl at least once, but not reliable enough to receive a finite score.
\Cref{tab:speedups} compares the training time speed-ups of all submission relative to the \baseline in their respective ruleset.
We investigated the sensitivity of leaderboard rankings and benchmark scores to changes in $\tau_{\text{max}}$, the upper limit of the performance profile and integration for the benchmark score.
As shown in \cref{fig:scores_max_tau}, rankings remain relatively stable for most submissions across different $\tau_{\text{max}}$ values.
With \cref{fig:perf_profiles_ignore_heldouts,fig:perf_profiles_qualification_set,tab:leaderboard_ignoring_wls} we explore how hypothetical rules changes would affect the competition results.

\begin{table}[!htb]
    \caption{\textbf{Speed-ups vs.~the baseline.} To compute the speed-up over the baseline, we first compute the geometric mean of the \wl runtimes relative to the runtimes of the baseline. Only fixed base \wls are considered. If a submission did not reach the target on a \wl, its runtime is imputed with the runtime budget, \ie assuming that the submission would have reached the target just after the cut-off. This is the best-case assumption for the submissions. We do not set runtimes to infinity, \eg because the corresponding held-out \wl was not trained successfully. The geometric mean is then expressed as a relative speed-up over the baseline, with positive numbers representing faster training.}
    \label{tab:speedups}
    \begin{subtable}[t]{.5\linewidth}
      \centering
        \caption{\textbf{External tuning}}
        \label{tab:speedups_et}
        \small
        {\renewcommand{\arraystretch}{1.25}
\setlength{\tabcolsep}{4pt}
\begin{tabularx}{0.85\textwidth}{XS[table-format=2.2]}
\toprule
\textbf{Submission} & \textbf{Speed-up} \\
\midrule
\shampoosubshort & +27.87\% \\
\sfadam & +26.60\% \\
\caspr & +24.33\% \\
\generalizedadam & +10.89\% \\
\cycliclr & +10.67\% \\
\lawaq & +7.98\% \\
\nadamp & +3.37\% \\
\amos & +1.46\% \\
\rowcolor{TUgray_light}\baseline & -0.00\% \\
\lawaema & -9.17\% \\
\sfprodigy & -15.68\% \\
\bottomrule
\end{tabularx}
}
     \end{subtable}\begin{subtable}[t]{.5\linewidth}
      \centering
        \caption{\textbf{Self-tuning}}
        \label{tab:speedups_st}
        \small
        {\renewcommand{\arraystretch}{1.25}
\setlength{\tabcolsep}{4pt}
\begin{tabularx}{0.85\textwidth}{XS[table-format=3.2]}
\toprule
\textbf{Submission} & \textbf{Speed-up} \\
\midrule
\sfadam & +7.76\% \\
\rowcolor{TUgray_light}\baseline & -0.00\% \\
\nadamwseq & -92.44\% \\
\sinvnum & -157.67\% \\
\sinv & -168.63\% \\
\adamg & -294.16\% \\
\bottomrule
\end{tabularx}
}
     \end{subtable} 
\end{table}

\begin{table}[!htb]
    \caption{\textbf{Submission benchmark scores and ranking when removing individual \wls.}
    Shown are the benchmark scores (S.) and leaderboard rankings (R.) of all external tuning (\cref{tab:leaderboard_ignoring_wls_et}) and self-tuning (\cref{tab:leaderboard_ignoring_wls_st}) submissions, when dropping specific \wls. For example, the last column reports the rankings of the submissions if we consider all \wls (including held-out \wls) \emph{except} the \wmt \wls. The ``Full'' columns show the scores and ranks when considering all \wls.}
    \label{tab:leaderboard_ignoring_wls}
    \begin{subtable}[c]{\linewidth}
      \centering
        \caption{\textbf{External tuning ruleset}}
        \label{tab:leaderboard_ignoring_wls_et}
        \scriptsize
	    {\renewcommand{\arraystretch}{1.25}
\setlength{\tabcolsep}{4pt}
\begin{tabularx}{0.99\textwidth}{@{}p{1.9cm}S[table-format=1.2]gS[table-format=1.2]gS[table-format=1.2]gS[table-format=1.2]gS[table-format=1.2]gS[table-format=1.2]gS[table-format=1.2]gS[table-format=1.2]gS[table-format=1.2]g@{}}
\toprule
 & \multicolumn{2}{c}{Full} & \multicolumn{2}{c}{\makecell{\textsc{Criteo}\\ \textsc{1TB}}} & \multicolumn{2}{c}{\fastmri} & \multicolumn{2}{c}{\resnet} & \multicolumn{2}{c}{\vit} & \multicolumn{2}{c}{\makecell{\textsc{Con-}\\ \textsc{former}}} & \multicolumn{2}{c}{\makecell{\textsc{Deep}\\ \textsc{Speech}}} & \multicolumn{2}{c}{\ogbg} & \multicolumn{2}{c}{\wmt} \\
 \cmidrule(lr){2-3} \cmidrule(lr){4-5} \cmidrule(lr){6-7} \cmidrule(lr){8-9} \cmidrule(lr){10-11} \cmidrule(lr){12-13} \cmidrule(lr){14-15} \cmidrule(lr){16-17} \cmidrule(lr){18-19}
 & \text{Score} & \text{Rank} & \text{S.} & \text{R.} & \text{S.} & \text{R.} & \text{S.} & \text{R.} & \text{S.} & \text{R.} & \text{S.} & \text{R.} & \text{S.} & \text{R.} & \text{S.} & \text{R.} & \text{S.} & \text{R.} \\
\midrule
\pytorch \textsc{Distr.} \newline \textsc{Shampoo} & 0.78 & 1 & 0.75 & 1 & 0.75 & 1 & 0.89 & 1 & 0.75 & 1 & 0.75 & 1 & 0.75 & 1 & 0.77 & 2 & 0.81 & 1 \\
\sfadam & 0.71 & 2 & 0.67 & 2 & 0.67 & 2 & 0.81 & 2 & 0.68 & 2 & 0.68 & 3 & 0.68 & 2 & 0.81 & 1 & 0.67 & 2 \\
\generalizedadam & 0.64 & 3 & 0.60 & 3 & 0.61 & 4 & 0.59 & 6 & 0.63 & 3 & 0.73 & 2 & 0.59 & 3 & 0.73 & 3 & 0.63 & 3 \\
\cycliclr & 0.63 & 4 & 0.58 & 4 & 0.62 & 3 & 0.72 & 3 & 0.62 & 4 & 0.59 & 4 & 0.59 & 4 & 0.72 & 4 & 0.60 & 4 \\
\nadamp & 0.59 & 5 & 0.54 & 6 & 0.57 & 5 & 0.68 & 4 & 0.58 & 5 & 0.55 & 5 & 0.53 & 5 & 0.68 & 5 & 0.60 & 5 \\
\baseline & 0.57 & 6 & 0.53 & 8 & 0.55 & 6 & 0.65 & 5 & 0.56 & 6 & 0.52 & 7 & 0.52 & 6 & 0.65 & 6 & 0.58 & 6 \\
\amos & 0.49 & 7 & 0.56 & 5 & 0.50 & 7 & 0.56 & 7 & 0.44 & 7 & 0.42 & 9 & 0.42 & 8 & 0.56 & 7 & 0.47 & 8 \\
\textsc{CASPR} \newline \textsc{Adaptive} & 0.47 & 8 & 0.54 & 7 & 0.40 & 8 & 0.54 & 8 & 0.41 & 8 & 0.54 & 6 & 0.41 & 9 & 0.40 & 8 & 0.54 & 7 \\
\lawaq & 0.37 & 9 & 0.42 & 9 & 0.32 & 9 & 0.42 & 9 & 0.31 & 9 & 0.42 & 8 & 0.42 & 7 & 0.33 & 10 & 0.31 & 10 \\
\lawaema & 0.34 & 10 & 0.25 & 10 & 0.31 & 10 & 0.39 & 10 & 0.28 & 10 & 0.39 & 10 & 0.39 & 10 & 0.39 & 9 & 0.32 & 9 \\
\sfprodigy & 0.00 & 11 & 0.00 & 11 & 0.00 & 11 & 0.00 & 11 & 0.00 & 11 & 0.00 & 11 & 0.00 & 11 & 0.00 & 11 & 0.00 & 11 \\
\bottomrule
\end{tabularx}
}
     \end{subtable}
    \par\bigskip \begin{subtable}[c]{\linewidth}
      \centering
        \caption{\textbf{Self-tuning ruleset}}
        \label{tab:leaderboard_ignoring_wls_st}
    	\scriptsize
    	{\renewcommand{\arraystretch}{1.25}
\setlength{\tabcolsep}{4pt}
\begin{tabularx}{0.99\textwidth}{@{}p{1.9cm}S[table-format=1.2]gS[table-format=1.2]gS[table-format=1.2]gS[table-format=1.2]gS[table-format=1.2]gS[table-format=1.2]gS[table-format=1.2]gS[table-format=1.2]gS[table-format=1.2]g@{}}
\toprule
 & \multicolumn{2}{c}{Full} & \multicolumn{2}{c}{\makecell{\textsc{Criteo}\\ \textsc{1TB}}} & \multicolumn{2}{c}{\fastmri} & \multicolumn{2}{c}{\resnet} & \multicolumn{2}{c}{\vit} & \multicolumn{2}{c}{\makecell{\textsc{Con-}\\ \textsc{former}}} & \multicolumn{2}{c}{\makecell{\textsc{Deep}\\ \textsc{Speech}}} & \multicolumn{2}{c}{\ogbg} & \multicolumn{2}{c}{\wmt} \\
 \cmidrule(lr){2-3} \cmidrule(lr){4-5} \cmidrule(lr){6-7} \cmidrule(lr){8-9} \cmidrule(lr){10-11} \cmidrule(lr){12-13} \cmidrule(lr){14-15} \cmidrule(lr){16-17} \cmidrule(lr){18-19}
& \text{Score} & \text{Rank} & \text{S.} & \text{R.} & \text{S.} & \text{R.} & \text{S.} & \text{R.} & \text{S.} & \text{R.} & \text{S.} & \text{R.} & \text{S.} & \text{R.} & \text{S.} & \text{R.} & \text{S.} & \text{R.} \\
\midrule
\sfadam & 0.85 & 1 & 0.83 & 1 & 0.83 & 1 & 0.98 & 1 & 0.83 & 1 & 0.83 & 1 & 0.85 & 1 & 0.83 & 1 & 0.84 & 1 \\
\baseline & 0.82 & 2 & 0.79 & 2 & 0.82 & 2 & 0.94 & 2 & 0.81 & 2 & 0.79 & 2 & 0.79 & 2 & 0.81 & 2 & 0.79 & 2 \\
\textsc{NadamW}\newline \textsc{Sequential} & 0.33 & 3 & 0.38 & 3 & 0.27 & 3 & 0.38 & 3 & 0.30 & 3 & 0.38 & 3 & 0.29 & 3 & 0.27 & 3 & 0.38 & 3 \\
\sinvnum & 0.14 & 4 & 0.16 & 4 & 0.12 & 4 & 0.16 & 4 & 0.16 & 4 & 0.16 & 4 & 0.13 & 4 & 0.16 & 4 & 0.08 & 4 \\
\sinv & 0.09 & 5 & 0.10 & 5 & 0.07 & 5 & 0.10 & 5 & 0.10 & 5 & 0.10 & 5 & 0.09 & 5 & 0.10 & 5 & 0.04 & 5 \\
\adamg & 0.00 & 6\phantom{1} & 0.00 & 6\phantom{1} & 0.00 & 6\phantom{1} & 0.00 & 6\phantom{1} & 0.00 & 6\phantom{1} & 0.00 & 6\phantom{1} & 0.00 & 6 \phantom{1}& 0.00 & 6\phantom{1} & 0.00 & 6\phantom{1} \\
\bottomrule
\end{tabularx}
}
     \end{subtable} 
\end{table}

\clearpage
\subsection{\algoperf workload wall-clock time comparison between \jax \& \pytorch}
\label{sec:timing_frameworks}
To measure the wall-clock time performance of \jax \& \pytorch \wl implementations, we estimate the submission times for an \adamw baseline training algorithm to train to target for each of the workloads on the \benchmarkinghardware competition system. The \wls support the same training batch sizes for this \adamw baseline across frameworks.

The submission time accumulates the wall-clock times of the \texttt{update\_params} and the \texttt{data\_selection} calls and excludes any time spent on logging and checkpointing (which is disabled during scoring runs anyway). During the first few steps of training, there may be some additional overhead in these calls resulting from cache warm-ups and compilation costs of the model and update code. An accurate estimate would be based on enough steps that any especially slow initial steps play only a small role in the average step time. We found that running for \num{20}\% of the step hint allowed us to estimate the equilibrium step time well. We then calculated the full submission time by extrapolating the submission time over the \num{20}\% step hint to the full step hint.

We initially performed this measurement when the workloads were complete in the sense that they were functionally equivalent. After upgrading the \jax \& \pytorch packages, changing the CUDA driver versions for the hardware configuration, and implementing various improvements and best practices, we then repeated the measurement to capture the final state of the \wls performance across frameworks.
The projected submission times of the \jax \& \pytorch \wl are presented in \cref{tab:timing_differences.tex}.

\begin{table}[!htb]
    \caption{\textbf{Projected submission times for \adamw baseline on \jax \& \pytorch workloads before and after \wl performance adjustments.} All submission times are in minutes. A positive difference indicates that our \jax implementation is faster than our \pytorch implementation.}
    \label{tab:timing_differences.tex}
    \footnotesize
    {\renewcommand{\arraystretch}{1.25}
\setlength{\tabcolsep}{4pt}
\hyphenpenalty=10000\exhyphenpenalty=10000
\begin{tabularx}{0.6\textwidth}{XS[table-format=4]>{\columncolor{TUgray_vlight}}S[table-format=4]S[table-format=4]>{\columncolor{TUgray_vlight}}S[table-format=4]S[table-format=2]>{\columncolor{TUgray_vlight}}S[table-format=2]}
\toprule
\textbf{Workload} & \multicolumn{2}{c}{\textbf{\jax}} & \multicolumn{2}{c}{\textbf{\pytorch}} & \multicolumn{2}{c}{\textbf{Difference}} \\
\cmidrule(l){2-3} \cmidrule(l){4-5} \cmidrule(l){6-7}
                  & \text{Before}          & \text{After}          & \text{Before}            & \text{After}            & \text{Before}              & \text{After}             \\ \midrule
\criteo           & 127             & 136            & 213               & 122              & 68\%                & -10\%             \\
\fastmri          & 148             & 145            & 163               & 163              & 10\%                & 12\%              \\
\resnet           & 1047            & 1063           & 1174              & 1135             & 12\%                & 7\%               \\
\vit              & 1290            & 1378           & 1260              & 1253             & -2\%                & -9\%              \\
\conformer        & 1689            & 1445           & 1755              & 1407             & 4\%                 & -3\%              \\
\deepspeech       & 1047            & 875            & 1229              & 967              & 17\%                & 10\%              \\
\ogbg             & 306             & 399            & 398               & 445              & 30\%                & 12\%              \\
\wmt              & 804             & 782            & 972               & 792              & 21\%                & 1\%               \\ \bottomrule
\end{tabularx}
}
     \centering
\end{table}

\end{document}